\documentclass[conference]{IEEEtran}
\usepackage{times}

\usepackage[numbers]{natbib}
\usepackage{multicol}
\let\labelindent\relax
\usepackage{enumitem}

\usepackage{graphicx}
\usepackage{float}
\usepackage{etoolbox}
\usepackage{caption}
\usepackage{multirow}
\usepackage{booktabs}
\usepackage{bm}
\usepackage{amsmath}

\usepackage[usenames,dvipsnames,table]{xcolor}
\definecolor{lightred}{rgb}{0.988, 0.294, 0.0823}
\definecolor{lightblue}{rgb}{0.11, 0.541, 0.752}
\definecolor{lightgreen}{rgb}{0.3, 0.8, 0.3}
\usepackage[bookmarks=true]{hyperref}
\hypersetup{
    colorlinks=true,
    linkcolor=MidnightBlue,
    filecolor=MidnightBlue,      
    urlcolor=MidnightBlue,
    citecolor=MidnightBlue,
}
\usepackage{cleveref}

\usepackage{todonotes}

\usepackage{amsmath}
\usepackage{todonotes}

\renewcommand{\vec}[1]{\bm{#1}}
\newcommand{\mat}[1]{\mathbf{#1}}

\begin{document}

\title{ViTaSCOPE: Visuo-tactile Implicit Representation for In-hand Pose and Extrinsic Contact Estimation}

\author{
    \authorblockN{
        \textbf{Jayjun Lee} \quad
        \textbf{Nima Fazeli}
    }
    \authorblockA{Robotics Department\\ University of Michigan}
    \vspace{1mm}
    \authorblockA{
        \textbf{\textcolor{MidnightBlue}{\url{https://jayjunlee.github.io/vitascope}}}
    }
}
\vspace{2mm}

\makeatletter
\apptocmd{\@maketitle}{
    \centering
    \includegraphics[width=0.85\textwidth, keepaspectratio=true]{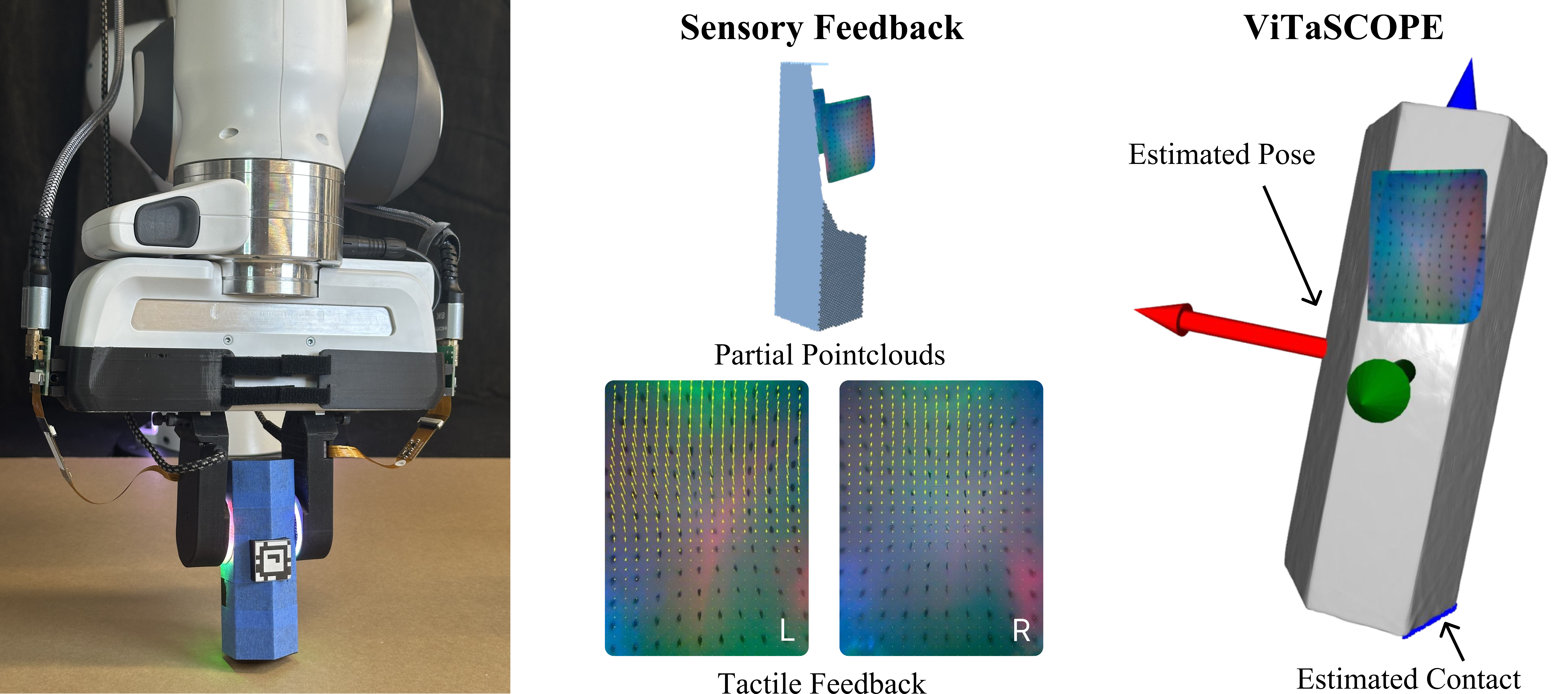}
    \captionof{figure}{\textbf{ViTaSCOPE: \underline{Vi}suo-\underline{Ta}ctile \underline{S}imultaneous \underline{C}ontact and \underline{O}bject \underline{P}ose \underline{E}stimation.} We present an object-centric neural implicit representation that enables simultaneous in-hand object pose and extrinsic contact estimation from vision and high-resolution tactile sensing. Our method is trained entirely in simulation and zero-shot transferred to the real world. In the example above, the tool grasped by the robot makes \textit{extrinsic contact} with the table. ViTaSCOPE is able to infer the in-hand object pose and register the extrinsic contact  (patch highlighted in blue) to the 3D geometry from partial real-world observations.}
    \label{fig:teaser_below_title}
}{}{}
\makeatother

\maketitle

\begin{abstract}

Mastering dexterous, contact-rich object manipulation demands precise estimation of both in-hand object poses and external contact locations—tasks particularly challenging due to partial and noisy observations. We present ViTaSCOPE: Visuo-Tactile Simultaneous Contact and Object Pose Estimation, an object-centric neural implicit representation that fuses vision and high-resolution tactile feedback. By representing objects as signed distance fields and distributed tactile feedback as neural shear fields, ViTaSCOPE accurately localizes objects and registers extrinsic contacts onto their 3D geometry as contact fields. Our method enables seamless reasoning over complementary visuo-tactile cues by leveraging simulation for scalable training and zero-shot transfers to the real-world by bridging the sim-to-real gap. We evaluate our method through comprehensive simulated and real-world experiments, demonstrating its capabilities in dexterous manipulation scenarios.
\end{abstract}

\section{Introduction}

Dexterous manipulation in contact-rich environments requires a seamless interplay of perception and control, particularly in unstructured settings where visual occlusions and unpredictable forces are the norm. From grasping a tool to executing precise interactions with the environment, the ability to reconstruct an object’s geometry, estimate its in-hand pose, and sense external contacts is critical for enabling effective manipulation. These capabilities are not merely enhancements; they are foundational for planning, predicting, and executing actions. Key to unlocking these capabilities is the ability to exploit the complementary nature of vision and touch.

However, achieving this level of perception remains challenging due to the inherent ambiguity of partial observations from vision or tactile sensing alone. Vision offers a global but often occluded view of the environment, while tactile sensing provides high-resolution, localized contact information that complements vision but is limited to regions of intrinsic contact. Bridging these modalities is essential for integrating global context with fine-grained contact details, which requires understanding not only the 3D geometry of objects but also the behavior of extrinsic contacts that induce shear forces on tactile sensor membranes.

To address these challenges, we propose ViTaSCOPE: a unified neural implicit representation for Visuo-Tactile Simultaneous extrinsic Contact and in-hand Object Pose Estimation. ViTaSCOPE introduces a novel framework that integrates vision and high-resolution tactile feedback to reconstruct object geometry, estimate in-hand object pose, and localize extrinsic contacts. Our approach offers the following key contributions:

\begin{itemize}[leftmargin=*] 
    \item We propose a contact-aware implicit representation that encodes object geometry as a signed distance field (SDF) and registers extrinsic contacts to the 3D geometry by conditioning on tactile shear field measured by vision-based tactile sensors. This enables seamless fusion of vision and tactile observations as an implicit function, enabling functionalities such as object geometry reconstruction, in-hand pose estimation, and extrinsic contact localization. Unlike traditional methods, our representation provides infinite resolution for object geometry and tactile shear fields, unlocking new capabilities for precision manipulation.
    \item ViTaSCOPE offers a modularized approach to integrating distributed tactile feedback for contact estimation where its inference process is interpretable.
    \item We scale up the dataset for extrinsic contact estimation using a penalty-based tactile model in GPU-accelerated simulation \cite{tacsl}. This approach enables robust sim-to-real transfer, leveraging the shear field tactile feedback for reliable performance compared to tactile RGB images. 
\end{itemize}

\section{Related Works}

\subsection{Simulation of Vision-based Tactile Sensors}
Vision-based tactile sensors such as DIGIT \cite{digit}, GelSight \cite{gelsight}, GelSlim \cite{gelslim3, gelslim4}, Soft Bubble \cite{softbubble}, and DIGIT360 \cite{digit360} provide high-resolution contact information through the observation of elastometric sensor membrane that deforms upon contact. The tactile sensors not only provide local texture and geometric information about the object from its contact patch, but also contain information about shear and normal force distributions during contact. These distributions describe pressure, shear, torsion, and slip along the tactile membrane. Recently, there has been advances in simulating these vision-based tactile sensors. TACTO \cite{tacto} can render realistic high-resolution tactile signatures as RGB-D images by simulating rigid-body contacts between the elastomer and a rigid object. TacSL \cite{tacsl} adopts the soft contact penalty-based tactile model \cite{tactilesim} with rigid bodies which allows interpenetration to simulate imaging process where contact force fields can be computed with GPU-accelerated Physics in Issac Gym \cite{isaacgym}. Taxim \cite{taxim} proposes an example-based method that combines optical simulation and marker motion field simulation that is calibrated with contact examples. DiffTactile \cite{difftactile} introduces a physics-based differentiable tactile simulation of contact force distribution and surface deformation using Finite-Element Method-based soft body model for the elastomers. M3L \cite{m3l} uses global visual image with shear force field images in MuJoCo \cite{mujoco} with its touch-grid sensor plugin. In this work, we use TacSL to simulate shear force field observations from extrinsic contact events during prehensile manipulation with tools.

\subsection{Implicit Object and Contact Representations}
Prior works on dense 3D geometry representations provide high-fidelity surface reconstructions for complex volumetric geometries. DeepSDF~\cite{deepsdf} introduces a learned continuous signed distance function (SDF) representation that enables high-quality shape representation, interpolation, and completion from partial and noisy 3D input data in canonical pose of the object. VIRDO~\cite{virdo, virdopp} extends these concepts by integrating multimodal feedback from vision and wrench from force/torque sensing to learn rich latent embeddings of contact locations and forces to predict tool deformations subject to extrinsic contacts. Building on these foundations, NDCF~\cite{ndcf} jointly models object deformations and extrinsic contact patches from similarly vision and wrench feedback using implicit representations, while ensuring that the predicted contacts lie on the object's surface. NCF~\cite{ncf} tracks extrinsic contact by an implicit function that maps a query point to a contact probability, which does not consider multi-modal inputs, only vision-based tactile inputs and assumes rigid, known object models during inference. NISP~\cite{nisp} uses physics-informed neural networks to solve inverse source problems such as identifying unknown source functions for detecting intrinsic and extrinsic contacts via an implicit representation with physics integration.

Despite these advancements, prior approaches~\cite{virdo, ndcf, ncf} assume rigid grasps or extends the kinematic chain that directly associate the object pose with the end-effector pose. This simplification ensures that real-world point cloud observations align with the object's canonical pose, removing the need for an additional explicit optimization of object pose in $\mathrm{SE}(3)$ transformation space~\cite{deepsdf} before model inference. In contrast, our work removes the assumption of rigid grasps, allowing for in-hand object translations and rotations to address real-world observations in the world frame rather than the canonical pose.

\subsection{Object Shape Reconstruction and Pose Estimation}
Accurate object pose estimation is essential for robotics applications such as manipulation, navigation, and human-robot interaction. Vision-based methods, coupled with multimodal sensing, have significantly advanced the precision and robustness of 6D pose estimation, particularly in dynamic and unstructured environments.

Vision-based methods have significantly advanced object shape reconstruction and pose estimation, leveraging deep learning and implicit representations. AlignSDF~\cite{alignsdf} combines SDF with parametric models to reconstruct hands and objects from monocular RGB images, disentangling pose and shape for improved accuracy. FoundationPose~\cite{foundationpose} introduces a generalizable model for 6D pose estimation across diverse datasets, while What's in Your Hands?~\cite{whatsinyourhand} focuses on reconstructing hand-held objects from single RGB images using hand articulation as a cue for inferring object shape.

Tactile-based methods~\cite{tac2pose, bubblecontactpatch, trackingobjectsruss} achieve pose estimation from contact RGB image to contact patch estimation then run matching against a precomputed dense set of simulated contact patches to do object pose estimation via geometric contact rendering from virtual cameras in simulation with access to 3D object models. \cite{bubblecontactpatch} proposes a model-based approach where they estimate the contact patches on highly compliant Soft Bubble sensors~\cite{softbubble} and estimates object pose with ICP using the obtained tactile pointclouds.~\cite{trackingobjectsruss} fuses pointclouds from vision and touch while focusing on the intrinsic contact geometries. In this paper, we exploit the functionalities of implicit functions to perform both in-hand pose and extrinsic contact estimation in a single framework by utilizing both visuo-tactile point clouds and shear fields.

The integration of vision and touch has further improved 3D shape reconstruction and pose estimation.~\cite{shaperecon, activeshaperecon} demonstrate how combining these modalities enhances shape reconstruction quality, with the latter actively selecting tactile readings for optimal accuracy. Similarly, some works~\cite{vihope, visuotactile6dposeest} leverage visuo-tactile data to complete partially observed geometries and estimate poses robustly under varying conditions. However,~\cite{vihope, visuotactile6dposeest} both operate only in pointclouds and trains for the purpose of 6D pose estimation. In our work, we learn an implicit representation that can be used for pose estimation as one of its properties while simultaneously being able to predict extrinsic contact locations on the object by reasoning about shear force fields from touch.

Prior works such as SCOPE~\cite{scope} and MultiSCOPE~\cite{multiscope} tackle the problem of simultaneous contact and object pose estimation (SCOPE), which leverages contact particle filters (CPF)~\cite{cpf}, propriceptive feedback, and force/torque sensing to jointly estimate the object pose and the location of contact. However, these methods do not make use of high-resolution tactile sensing and require \textit{a priori} known and fixed object representations. There is currently no way to allow these methods to learn models online.

\subsection{Extrinsic Contact Estimation and Control}
Extrinsic contact sensing is an important yet challenging problem in contact-rich robotic manipulation due to the occlusions, indirect transmission of contact force/torque and the uncertainties in the object’s geometry, stiffness, and pose. Im2Contact~\cite{im2contact} proposes a sim-to-real method that does not rely on additional sensor modality such as tactile or force/torque sensing to achieve zero-shot visual extrinsic contact estimation by using object cropping heuristics around the EE and optical flow that provides temporal information. Another work~\cite{vacontact} leverages active-audio for predicting extrinsic contacts via visual-auditory feedback zero-shot in the real world using sim-to-real transfer.~\cite{im2contact, vacontact} both predict contacts in 2D depth images, which loses key depth information in the real world. Typically, visual extrinsic contact estimation suffers from a lack of 3D priors of object geometries and occlusions, which is often complemented by an additional sensor modality to capture a more direct contact cues at higher resolution i.e. audio~\cite{vacontact} or tactile sensing. 

Other prior works utilize implicit representations~\cite{virdo, ncf, ndcf} to jointly reason and estimate contact geometries on rigid and deformable objects. NCF-v2 \cite{ncfv2} learns a policy that can act while sensing and reasoning about extrinsic contacts with neural contact fields (NCF) but is limited to fixed or known in-hand poses~\cite{ncf}.

Dexterous regulation and control of extrinsic contacts are another set of important skills in robotic manipulation. \citet{ollercorl} decouples the interaction dynamics from the tactile observation model to control a grasped object for tasks such as drawing with a pen and pivoting spatulas through extrinsic contacts. \cite{ollerrss} enables pivoting of an object using a grasped object with bubble sensors via extrinsic contact mode control. \cite{ext_contact_sensing_w_rel_motion} proposes an approach to infer the contact location from tactile shear observations from small motions without requiring the knowledge of geometry but is limited to fixed point or line contact modes. \cite{tactile_dense_packing, tactile_force_est_iFEM, simult_tact_est_and_cont_of_ext_contact, texterity} uses GelSlim tactile sensors to sense extrinsic contacts and estimate force distributions on the sensor membrane for simultaneous tactile estimation and control for extrinsic dexterity for various tasks such as insertion. In this paper, we introduce a scalable sim-to-real method for learning object representations that can reason about in-hand object poses and tactile cues for perceiving extrinsic contacts.

\section{Methodology}
\subsection{Problem Formulation and Assumptions}

\begin{figure*}[t]
    \centering
    \includegraphics[width=\linewidth]{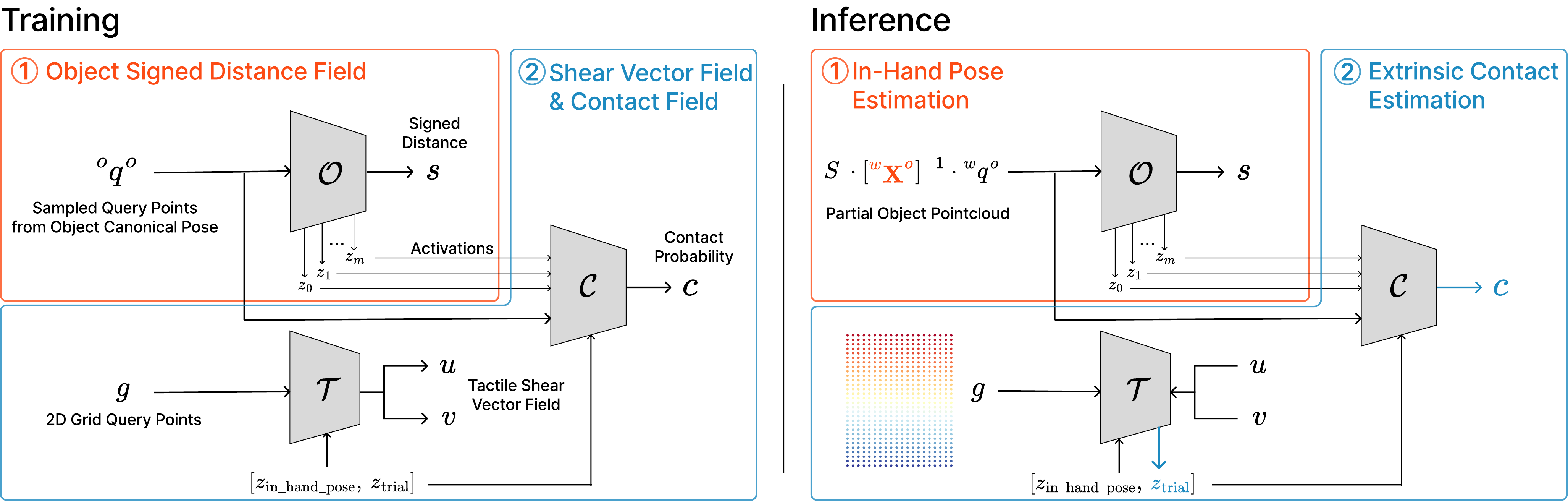}
    \caption{\textbf{ViTaSCOPE training and inference.} ViTaSCOPE is composed of three modules: Object module $\mathcal{O}$, Tactile module $\mathcal{T}$, and Contact module $\mathcal{C}$. We represent the object geometry as a signed distance function (SDF) $s$ using the object module. This representation first enables joint reasoning over visuo-tactile pointcloud observations to perform \textcolor{lightred}{in-hand pose estimation} using inference via optimization. The tactile module represents the shear vector field in the local coordinates $(s_x,s_y)$ on a 2D grid of tactile sensing plane to capture the extrinsic contact configuration conditioned on the in-hand pose $z_\xi$ via the latent trial code $z_\psi$. At inference time, the \textcolor{lightblue}{extrinsic contact trial code} is inferred given the observed shear vector field over the grid points on the distributed left and right tactile sensors. We train the contact module to \textcolor{lightblue}{predict the contact probability $c$} by representing it as a contact field conditioned on layer-wise activations $z_\mathcal{O}=z_{0:m}$ of the object module and the tactile trial code $z_\psi$.}
    \label{fig:framework}
\end{figure*}
We propose ViTaSCOPE, a neural implicit representation that can fuse complementary visuo-tactile information to reason over in-hand pose of a grasped object and its extrinsic contact formations from visuo-tactile feedback as outlined in Fig. \ref{fig:teaser_below_title}. The key insight of our method is to jointly predict the object geometry and the contact geometry on the object surface while using shear fields observations to characterize the extrinsic contacts. One challenge with this approach is being able to collect large scale contact interactions with tactile shear data and extrinsic contact labels. Therefore, we opt for training with simulated data and bridging the sim-to-real gap effectively. Moreover, prior works \cite{virdo, ndcf, ncf, ncfv2} assumes known object poses and rigid grasps such that they have access to the object surface query points in its canonical pose. In this paper, we do not assume access to object pose and instead infer the object pose from visuo-tactile point clouds in the world frame, which is considered a raw scene observation, to retrieve the query points in the object canonical pose on which the neural fields are trained on. We choose to represent the object geometry, shear, and contact patch as neural fields. By learning the object geometry and contact patch jointly, we can enforce physical priors during training and inference, ensuring that contacts lie on the surface of the object and associate geometrical information with contact prediction. 

Our method is designed to incorporate visuo-tactile feedback from RGB-D cameras and vision-based tactile sensors. We assume access to a partial visual pointcloud $\tilde{P}_v \in \mathrm{R}^{N_v\times3}$ of the segmented object, tactile pointcloud  $\tilde{P}_t \in \mathrm{R}^{N_t\times3}$ that form $\tilde{P}=\{\tilde{P}_t,\ \tilde{P}_v \}$ together. We also assume access to the tactile shear vector field $\vec \phi \in \mathrm{R}^2$ and a tactile grid points on the sensor membrane $\vec g \in G = \{ (s_x,s_y) \ | \ s_x \in \left[ s_{x_{\text{min}}}, s_{x_\text{max}} \right], s_x \in \left[ s_{y_{\text{min}}}, s_{y_\text{max}} \right]  \}$. Then, given a 3D query point $\vec q \in \mathrm{R}^3$, we predict its SDF value $s \in \mathrm{R}$ and the likelihood that it is in contact $c \in [0, 1]$ and for tactile grid 2D query point $\vec g \in \mathrm{R}^2$ we predict the shear vector field $\vec \phi$ where its values correspond to $[u,v]^T \in \mathrm{R}^2$:

\begin{equation}
    (s,c,\phi) = f(\tilde{P},\vec q,\vec g)
\end{equation}

The object surface is given as the zero-level set of the signed distance field, which can be recovered through Marching Cubes \cite{marchingcubes} algorithm or ray tracing methods and can be easily used to generate a point cloud or a mesh.
\begin{equation}
    S = \{ \vec q \mid (s=0,c,\vec \phi) = f(\tilde{P},\vec q,\vec g) \}
\end{equation}

The contact patch geometry is given by the intersection of the object surface with points classified as in contact, where $\epsilon$ is the binary classification threshold.
\begin{equation}
    \label{Eqn:3}
    C = S \cap \{ \vec q \mid (s=0,c>\epsilon,\vec \phi) = f(\tilde{P},\vec q,\vec g) \}
\end{equation}

\subsection{Architecture}

ViTaSCOPE is designed to infer object geometry, in-hand pose, and extrinsic contacts by reasoning over visuo-tactile observations. To achieve this, we adopt an implicit function-based approach using auto-decoders to learn geometric, tactile, and contact representations. The architecture consists of three core modules (Fig.~\ref{fig:framework}): (i) \textbf{Object Module ($\mathcal{O}$)} whicn represents the object’s 3D geometry as a signed distance field (SDF); (ii) \textbf{Tactile Module ($\mathcal{T}$)} which models the tactile shear field induced by extrinsic contact interactions;
and (iii) \textbf{Contact Module ($\mathcal{C}$)} which estimates the likelihood of extrinsic contact per query point given object pose, geometry, and tactile observations. 

\vspace{2mm} \noindent
\textbf{Object Geometry Representation:}
We define the object’s shape as a continuous signed distance function (SDF), parameterized by a neural network $\mathcal{O}$. Given a query point $q \in \mathrm{R}^3$ in the object’s canonical frame, the network predicts its signed distance $s$, allowing for reconstruction of the object’s surface:
\begin{equation}
    s = \mathcal{O}(\vec q)
\end{equation}
This formulation enables a compact, high-resolution representation of object geometry without requiring explicit mesh storage  \cite{deepsdf, virdo, ndcf}.

\vspace{2mm} \noindent
\textbf{Tactile Shear Field Representation:}
To model extrinsic contact interactions, we introduce a \textbf{tactile shear field} represented as an implicit function. The \textbf{Tactile Module} ($\mathcal{T}$) takes as input a 2D query point $\vec{g} \in \mathrm{R}^2$ on the tactile sensor’s sensing plane and predicts the local shear displacement field:
\begin{equation}
    [u, v]^T = \mathcal{T}(\vec g \mid \vec{\xi}, \vec \psi)
\end{equation}
where $\vec \xi$ is a latent representation of the in-hand pose and $\vec \psi$ is a trial code characterizing the contact configuration. $\mathcal{T}$ is conditioned on both the in-hand pose and the trial codes as the same extrinsic contact interaction at varying in-hand object poses can induce different shear feedback. The trial code is learned during training and optimized at inference to explain the observed shear field. To support reasoning over a set of distributed tactile sensors, we instantiate a tactile module for each tactile sensor and concatenate the trial codes from each sensor along with the in-hand object pose code.

\vspace{2mm} \noindent
\textbf{Extrinsic Contact Prediction:}
Extrinsic contacts are inferred through the \textbf{Contact Module} ($\mathcal{C}$), which estimates the probability of contact at a given 3D query point. To incorporate object geometry, we condition $\mathcal{C}$ on the activations from $\mathcal{O}$, allowing it to use the object's geometric features:
\begin{equation}
    c = C(\vec q \oplus \vec{z}_\mathcal{O}, \mathcal{T}(\vec g \mid \vec \xi, \vec \psi) \mid \vec \xi, \vec \psi)
\end{equation}
where $\vec{z}_\mathcal{O}$ represents the intermediate activations from $\mathcal{O}$. This formulation ensures that extrinsic contacts are predicted in a geometry-aware manner, capturing interactions between the grasped object and the environment.

\vspace{2mm} \noindent
\textbf{Unified Implicit Representation:}
ViTaSCOPE jointly models object shape, in-hand pose, and contact information within a single framework (Fig.~\ref{fig:framework}). Unlike traditional methods that treat pose estimation and contact localization as separate tasks, our architecture unifies these functions through implicit representations. This allows for: (i) \textbf{High-resolution geometry reconstruction} without explicit meshes; (ii) \textbf{Generalization across different contact interactions} by learning latent contact embeddings; and (iii) \textbf{Seamless integration of visuo-tactile feedback}, enabling robust state estimation.

Each module is parameterized by a neural network, whose specific architectural implementation details and hyperparameters are provided in Section \ref{sec:impl}.

\subsection{Training}
Training ViTaSCOPE is composed of two stages. The object module $\mathcal{O}$ is first pre-trained to learn the object geometry without shear field or contact predictions. Then, the object module is combined with other tactile and contact modules and trained to jointly learn shear and contact fields.

The object module $\mathcal{O}$ is trained to fit the implicit function to our object shape. We sample off-, near-, and on-surface query points $\vec q_i$ with signed distance $s_i$ and surface normal $\vec n_i$ labels from the object mesh in its normalized canonical pose that fits into a unit sphere. Given this dataset of query samples $\Omega_i = \{ \vec q_j, s_j, \vec n_j \}_i$ of an object, we pretrain $\mathcal{O}$ with the following SDF loss.
\begin{equation}
    \mathcal{L}_\text{sdf} = \frac{1}{|\Omega|} \sum_{j=1}^{|\Omega|} |\mathcal{O}(\vec q_j) - s_j| + \lambda \frac{1}{|\Omega_s|} \sum_{j=1}^{|\Omega_s|}(1-\langle \nabla \mathcal{O}(\vec q_j), \vec n_j^* \rangle)
\end{equation}
where $\Omega_s$ is a subset of sampled surface points and $\lambda$ balances the loss weights from the SDF error and the normal alignment error. By learning an implicit function of SDF, the predicted surface normal can be computed by differentiating the SDF output w.r.t $\vec{q}_j$ as $\nabla \mathcal{O}(\vec{q}_j) = \partial \mathcal{O}(\vec{q}_j)/ \partial \vec{q}_j$ via backpropagation. The object module weights are frozen after pretraining.

To train the entire model, we collect a dataset of contact interactions  $\mathcal{D} = \{ (\Omega_i, \Phi_i, \vec \psi_i) \}_{i=1}^N$ where $\Omega_i= \{ \vec{q}_j, s_j^*, \vec{n}_j^*, c_j^* \}_i$, $\Phi_i = \{ \vec g_j, \left[u_j^*, v_j^* \right]^T \}_i$ is the shear field, and $\vec \psi_i$ is the generated trial code. We compute the training loss as the following. 
\begin{equation}
    \mathcal{L}_\text{train} = \lambda_\text{shear} \mathcal{L}_\text{shear} + \lambda_\text{emb} \mathcal{L}_\text{emb} + \lambda_\text{hyper} \mathcal{L}_\text{hyper} + \lambda_\text{contact} \mathcal{L}_\text{contact} 
\end{equation}
\begin{equation}
    \mathcal{L}_\text{shear} = \frac{1}{|\Phi_i|} \sum_{j=1}^{|\Phi_i|} |\vec \phi(\vec g_j) - \left[ u_j^*, v_j^* \right]^T |
\end{equation}

where $\vec \phi(g_j) = \left[ \hat{u}_j, \hat{v}_j \right]^T$.

\begin{equation}
    \mathcal{L}_\text{embedding} = ||\vec \psi||_2^2
\end{equation}


\begin{equation}
    \mathcal{L}_{\text{contact}} = \frac{1}{|\Omega_{i,S}|} \sum_{j=1}^{|\Omega_{i,S}|} \text{BCE}(C(\vec q \oplus \vec{z}_\mathcal{O} | \cdot), c^*_j)
\end{equation}

\subsection{Inference}

At inference time, our model has access to a partial visual pointcloud and a tactile pointcloud from left and right tactile sensors in the world frame as well as the shear displacement field obtained via tracking the markers from vision-based tactile sensing. We first perform inference to find the object pose. Then, we infer the trial code based on shear field observations to predict contact registered to the 3D geometry of the object.

\vspace{2mm} \noindent
\textbf{Pose Estimation:} The visual observation ${}^w \vec q^\text{visual}_s$ and tactile observation ${}^w \vec q^\text{tactile}_s$ form the input surface query points ${}^w \vec q^o_s$ in the world frame. We first initialize the in-hand SE(3) object pose as the SE(2) constrained pose of the end-effector parameterized by $(x, z, \theta)$.
\begin{equation*}
    {}^w \mat X^\text{o}_\text{SE(2)} = {}^w \mat X^\text{ee} \ \cdot \ \text{Proj}_{\mathrm{SE}(2)}(\left[ {}^w \mat X^\text{ee} \right]^{-1} \ \cdot \ \left[ {}^w \mat X^\text{o}_\text{SE(3)} \right])
\end{equation*}
where $\Omega_s = \{ \vec q_{v_s}, \vec q_{t_s} \in \mathrm{R}^3 \}$, ${}^w \mat X^o \in \mathrm{SE}(2)$, and $\text{Proj}_{\mathrm{SE}(2)}$ projects the result of its argument (an element in $\mathrm{SE}(3)$) to $\mathrm{SE}(2)$. To compute the object pose, we minimize the following surface signed distance loss through the object module where the losses from vision and touch are equally weighted due to the stochastic nature of the inference process:
\begin{equation}
\underset{{}^w \mat X^\text{o}}{\mathrm{argmin}} \ \frac{1}{|\Omega_s|} \sum_{\Omega_s} \left| \mathcal{O}\left(S \cdot [^w \mat X^o]^{-1} \cdot {}^w\vec q^o_s; z_o \right) \right|
\end{equation}
where $S$ is a scale factor for normalizing the geometries into a unit sphere.

We use gradient descent through our model to minimize the loss. This method is often referred to as inference via optimization \citep{virdopp,ndcf} since the variable of interest is derived through the optimization program. 

\vspace{2mm} \noindent
\textbf{Extrinsic Contact Trial Code Inference:} Next, we optimize the trial code $\vec \psi$ in the same manner as pose estimation such that our tactile module $\mathcal{T}$ matches the observed shear field. We run inference via optimization in Eq.~\ref{eqn:trial_code_inf}. Our key insight is that shear fields can capture rich information about the extrinsic contact interaction through the force transmission, subject to in-hand object pose $\vec \xi$. By optimizing for the trial code that can best explain the observed shear field induced by extrinsic contacts per object geometry and the in-hand pose, we can extract meaningful information about the contact configuration as a latent code $\vec \psi$.
\begin{equation}
    \underset{\vec \psi}{\mathrm{argmin}} \ \frac{1}{|\vec \phi_i|} \sum_{j=1}^{|\vec \phi_i|} \left| \mathcal{T}(\vec g_j \mid \vec \xi, \vec \psi) - [u_j, v_j]^\top \right|
    \label{eqn:trial_code_inf}
\end{equation}
where $[u_j, v_j]^\top$ is the observed shear field.

\vspace{2mm} \noindent
\textbf{Contact Prediction:} Once the trial code is inferred, we run inference on the full model to predict the contact probability for each query points on the object surface. This is equivalent to evaluating our model at surface query points and reading the value of the contact field. The contact patch geometry can be retrieved as the intersection of points that are above the contact threshold and belong to the surface of the object as in Eq. \ref{Eqn:3}.

\section{Implementation}
\label{sec:impl}
\subsection{Model}

The tactile module $\mathcal{T}$ and contact module $\mathcal{C}$ are implemented as hypernetworks, where conditioning latent codes are employed to predict the weights of a secondary network. We instantiate two tactile modules corresponding to the distributed left and right tactile sensors. This network then processes the query point and outputs the corresponding field value, aligning with recent advancements in neural implicit methods that emphasize hypernetworks as the optimal approach for conditioning outputs \cite{virdopp, ndcf}.

\begin{equation*}
    [u, v]^T = \mathcal{T}(\vec g \mid H_\mathcal{T}(\vec{\xi}, \vec \psi))
\end{equation*}
\begin{equation*}
    c = C(\vec q \oplus \vec{z}_\mathcal{O}, \mathcal{T}(\vec g \mid \vec \xi, \vec \psi) \mid H_\mathcal{C}(\vec \xi, \vec \psi))
\end{equation*}

The hypernetworks $H_\mathcal{O}$, $H_\mathcal{T}$, and $H_\mathcal{C}$ predict the weights of the MLP of each module. We regularize the predicted weights as the following. The MLP of $\mathcal{O}$, $\mathcal{T}$, and $\mathcal{C}$ are implemented with two hidden layers of size 256. We set the latent code size of pose $\vec \xi$ as 3 for the $\mathrm{SE}(2)$ in-hand object pose $(x,z,\theta)$ and extrinsic contact $\vec \psi$ as 12 (left and right tactile sensors concatenated).
\begin{align*}
    \mathcal{L}_{\text{hyper}} = \lambda_\text{hyper} \Bigg( 
    &\frac{1}{|H_\mathcal{T}{(\vec \xi, \vec \psi})}| \|H_\mathcal{T}(\vec \xi, \vec \psi)\|_2^2 \\
    &+ \frac{1}{|H_\mathcal{C}{(\vec \xi, \vec \psi})}| \|H_\mathcal{C}(\vec \xi, \vec \psi)\|_2^2 \Bigg)
\end{align*}
$\lambda_\text{hyper}$ is a weighting on the regularization loss. 

\subsection{Training and Inference Details}
ViTaSCOPE training is split into two stages. First, the object module is pretrained for 50000 epochs using the Adam optimizer with the learning rate of $1e-5$ and a normal loss weighting of $\lambda_\text{normal}$ of 0.01. Then, with the frozen weights of the object module, ViTaSCOPE is trained using the Adam optimizer with the learning rate of $1e-5$. The latent codes $\psi$ are initialized from the zero mean Gaussian with the standard deviation of 0.1. We use $\lambda_\text{shear}=0.1$, $\lambda_\text{embedding}=0.2$, $\lambda_\text{hyper}=25.0$ and $\lambda_\text{contact}=2.0$ and train for 20 epochs.

\vspace{1mm}
ViTaSCOPE inference for in-hand pose estimation uses the Adam optimizer with the learning rate of $5e-3$ and a learning rate scheduler with a minimum learning rate of $1e-4$. The $\mathrm{SE}(2)$ in-hand pose estimate is initialized at the $\mathrm{SE}(2)$ EE pose that is projected onto the x-z plane of the EE frame. Once an estimate of the in-hand pose is obtained, we infer the extrinsic contact trial code using the Adam optimizer with the learning rate of $3e-2$. We initialize the in-hand pose estimate as the EE pose to accelerate the inference process.

\subsection{Data Generation}
\label{sec:data_gen}
\begin{figure}[t]
    \centering
    \includegraphics[width=\linewidth]{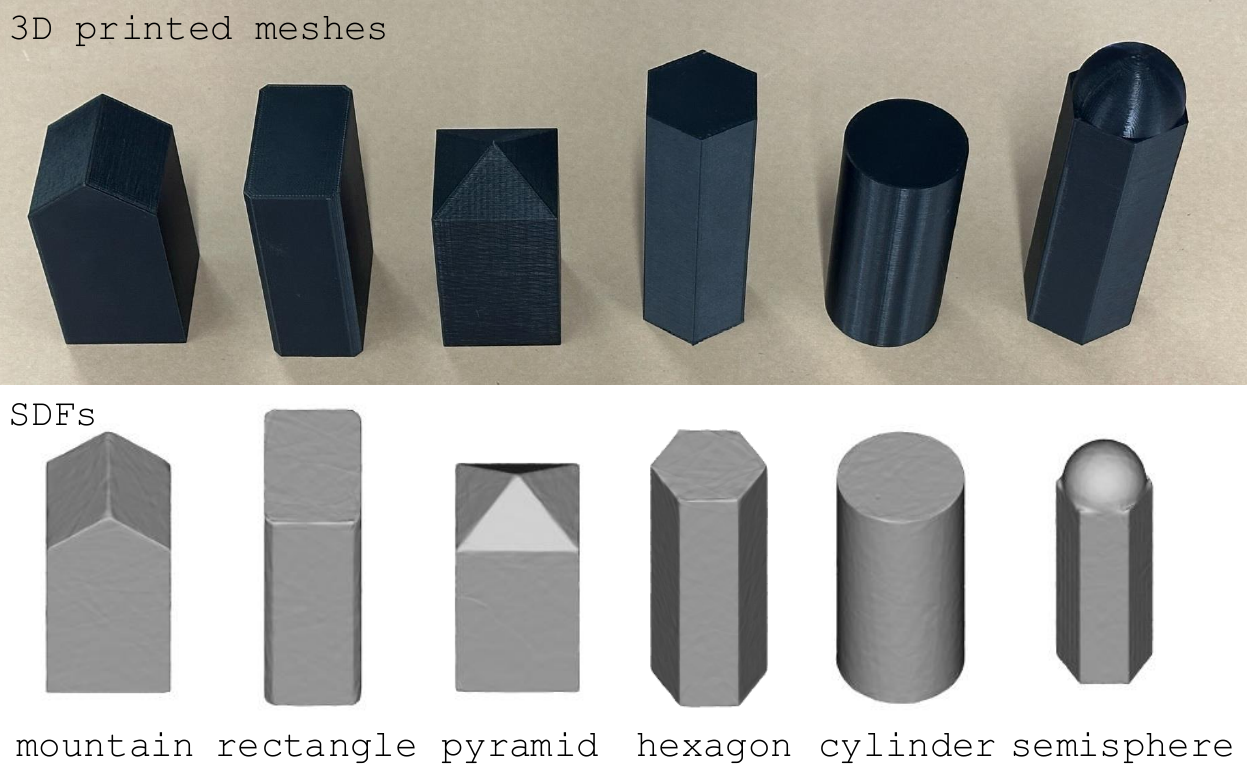}
    \caption{\textbf{3D Printed Tool Geometries and SDFs.} We 3D print six geometries for evaluation. The tool tip making contact with the environment is facing upwards. The number and complexity of possible extrinsic contact modes (i.e. point, line, patch) increases from left to right. Sample mesh reconstructions of learned SDF are shown in the bottom row.}
    \label{fig:tool_shapes}
    \vspace{-5mm}
\end{figure}

\textbf{Simulation:}
Model training and evaluation of ViTaSCOPE requires various multi-sensory data which can be challenging to collect at scale in the real world. To address this, we instead propose a sim-to-real approach where we leverage simulation to generate data to train the three modules ($\mathcal{O, T, C}$).
Once trained, we generate novel scene observations to evaluate on downstream in-hand pose estimation and extrinsic contact prediction tasks.

Training the \underline{Object Module} $\mathcal{O}$ requires a 3D mesh model of the object, which is first zero-centered and scaled by a factor $S$ to fit the mesh into a unit sphere ($r=1$). We sample 5000 off-surface query points, 15000 on-surface query points, and 20000 near-surface query points on this normalized mesh in its \textit{canonical pose} paired with ground truth signed distance and surface normals. We denote this as the object-centric canonical frame or pose, where the object module takes in query points defined in this object canonical frame. In this paper, we 3D print 6 tool geometries with varying tool-tip complexity as shown in Fig. \ref{fig:tool_shapes} for training and testing. We assume access to the mesh models of these geometries for training purposes. Fig. \ref{fig:tool_shapes} also visualizes sample meshes generated using the pretrained object module.

\begin{figure}[t!]
    \centering
    \includegraphics[width=\linewidth]{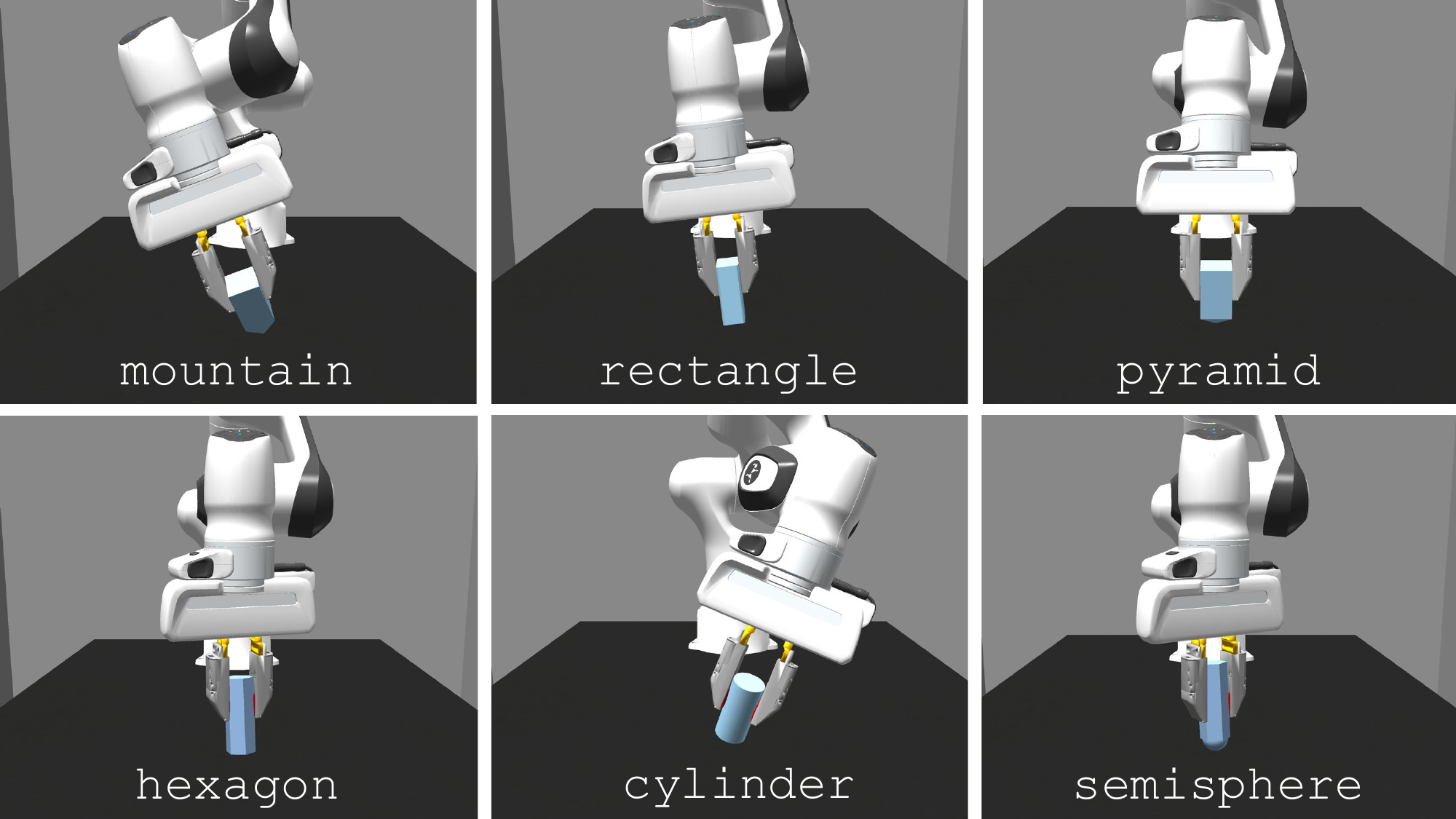}
    \caption{\textbf{Simulation Environment for Visuo-tactile Data Generation.} A visualization of simulated environments in Issac Gym and sampled extrinsic contact interactions of six different tool geometries with TacSL tactile sensor simulation.}
    \label{fig:simulation_setup}
    \vspace{-2mm}
\end{figure}

We use Isaac Gym \cite{isaacgym} coupled with TacSL's \cite{tacsl} vision-based tactile sensing simulation capabilities to generate a large dataset of contact locations, object poses, and visuo-tactile feedback. We split this dataset into training and test sets for training and inference. Fig. \ref{fig:simulation_setup} visualizes 6 sample scenes for each tool and a sampled contact interaction. We place eight cameras spaced evenly around the EE pose during contact to capture the partial point clouds of the target tool object. Using TacSL's implementation of the penalty-based tactile simulation, we collect simulated visuo-tactile feedback. 

The simulation data generation pipeline works as follows: First, we use the Franka Emika Panda robot with GelSight R1.5 vision-based tactile sensors on the grippers to grasp the tool at randomized pitch angle of the EE. Next, the robot lifts the object and pushes down on to the table at a random angle as visualized in Fig.~\ref{fig:simulation_setup}. This effectively generates diverse grasp poses of the tool and varying contact modes on the tool surface to induce a diverse distribution of tactile feedback.


The \underline{Tactile Module} $\mathcal{T}$ learns to predict a field of shear displacements. We first generate a uniform 2D grid of tactile points of shape $\left[ 30, 20 \right]$ with $ \left[ s_{x_\text{min}}, s_{x_\text{max}} \right] = \left[ -0.008, 0.008 \right] $ and $ \left[ s_{y_\text{min}}, s_{y_\text{max}} \right] = \left[ -0.01215, 0.01215 \right] $, which is scaled and normalized to fit into a square grid of length 1. For TacSL simulation, we set the compliance of the elastomer as 1500. This effectively maps each 2D query point to its corresponding shear displacement vectors from simulation. These shear vectors are normalized to unit vectors for those above a magnitude threshold of $1e-8$ to mitigate numerical instabilities and to bridge the sim-to-real gap caused by mismatches between simulated and real-world shear vectors. 

As Isaac Gym does not allow the users to access the contact locations directly, we replicate the scene in \texttt{open3d} and estimate the contact location by thresholding on object-table penetration, which effectively generates soft-contact points. For each contact scene, we collect 1000 contact points to train the \underline{Contact Module} $\mathcal{C}$. At training time, we normalize the individual shear displacement vectors and augment it by adding random Gaussian noises $\mathcal{N}(0,0.1)$ along the local $(s_x,s_y)$ tactile frames to account for real-world noises and the sim-to-real gap of the gel elastomer deformation dynamics of the tactile sensor.

For each tool contact simulation, we sample 500 contact interactions per tool, resulting in 3000 in total for six tools. For the simulated testing data, we generate 50 interactions per tool, resulting in 300 interactions in total.

\vspace{2mm}
\textbf{Real World:}
We outline the pipeline for collecting the real-world test data that takes an equivalent form of the simulated data. For the purpose of evaluation, we collect the scene observations and ground truth object poses. For the real world setup, we use the 7DoF Franka Emika Panda arm with a pair of Gelslim 4.0 vision-based tactile sensors. We set up a single front view Realsense D435 RGB-D camera with known extrinsics to obtain the scene point clouds in the world frame. We attach AprilTags to the tool geometries to track the object poses. Segmented tool point clouds are obtained by using Grounded SAM v2 \cite{gsam} open-vocabulary segmentation model. For the tactile point cloud, contact patch points on the sensor membrane is de-projected into the 3D scene. Given this perception pipeline, we command a similar robot motion as in simulation to collect various extrinsic contact interactions at randomized angles and in-hand pose. We collect 10 interactions per geometry, resulting in 60 trials of real world test data. 

\section{Experiments and Results}

\begin{table*}[t]
  \centering
  \begin{tabular}{l c ccc ccc}
    \toprule
    \multicolumn{1}{c}{} & \multicolumn{1}{c}{} & \multicolumn{3}{c}{Simulation} & \multicolumn{3}{c}{Real world} \\
    \cmidrule(lr){3-5} \cmidrule(lr){6-8}
    Geometries & \shortstack{SDF CD $\downarrow$ [$\text{m}^2$]} & \multicolumn{2}{c}{Pose Est.} & Ext. Contact Patch & \multicolumn{2}{c}{Pose Est.} & Ext. Contact Patch \\
    \cmidrule(lr){3-4} \cmidrule(lr){5-5} \cmidrule(lr){6-7} \cmidrule(lr){8-8}
             &  & Trans. err [mm] & Rot. err [deg] & CD [$\text{m}^2$] & Trans. err [mm] & Rot. err [deg] & CD [$\text{m}^2$] \\
    \midrule
    \texttt{mountain}   & 0.072  & 3.639  & 2.739  & 0.0015  & 3.110 & 1.286 & 0.0394 \\
    \texttt{rectangle}  & 0.057  & 2.504  & 4.833  & 0.2724  & 2.162 & 0.210 & 0.2216 \\
    \texttt{pyramid}    & 0.065  & 3.567  & 0.703  & 0.0676  & 3.203 & 1.487  & 0.0565 \\
    \texttt{hex}        & 0.041  & 3.415  & 0.872  & 0.0841  & 2.286 & 0.790 & 0.0596 \\
    \texttt{cylinder}   & 0.039  & 2.924  & 1.660  & 0.1534 & 3.893 & 1.011 & 0.1716 \\
    \texttt{semisphere} & 0.040  & 8.571  & 0.371  & 0.0017  & 9.637 & 0.657 & 0.0031 \\
    \texttt{average}    & 0.052  & 4.103  & 1.863  & 0.0968  & 4.049 & 0.907 & 0.0920 \\
    \bottomrule
  \end{tabular}
  \caption{\textbf{Results on Geometry Reconstruction, In-hand Pose Estimation, and Extrinsic Contact Prediction in Simulation and Real-world.} CD results are based on normalized geometries.}
  \label{tab:main_results}
\end{table*}
 

\begin{table*}[t]
  \centering
  \begin{tabular}{l cc cc cc cc cc cc}
    \toprule
    \multicolumn{1}{c}{} & \multicolumn{6}{c}{ICP} & \multicolumn{6}{c}{ViTaSCOPE (Ours)} \\
    \cmidrule(lr){2-7} \cmidrule(lr){8-13}
    Geometries & \multicolumn{2}{c}{Vision} & \multicolumn{2}{c}{Tactile} & \multicolumn{2}{c}{Vision+Tactile} & \multicolumn{2}{c}{Vision} & \multicolumn{2}{c}{Tactile} & \multicolumn{2}{c}{Vision+Tactile} \\
    \cmidrule(lr){2-3} \cmidrule(lr){4-5} \cmidrule(lr){6-7} \cmidrule(lr){8-9} \cmidrule(lr){10-11} \cmidrule(lr){12-13}
    & Trans. err [mm] & Rot. err [deg] & Trans. & Rot. & Trans. & Rot. & Trans. & Rot. & Trans. & Rot. & Trans. & Rot. \\
    \midrule
    \texttt{mountain}   & 81.4 & 134.8 & 81.6 & 70.9  & 81.8 & 49.22 & 16.6 & 9.96 & 16.5 & 21.4 & \textbf{3.11} & \textbf{1.29} \\
    \texttt{rectangle}  & 17.5 & 82.05 & 21.9 & 102.1 & 4.68 & 69.1 & \textbf{1.91} & 0.27 & 26.7 & 10.3 & \textbf{2.16} & \textbf{0.21} \\
    \texttt{pyramid}    & 129.2 & 85.4 & 119.6 & 160.9 & 122.7 & 58.4 & 3.35 & 1.51 & 26.4 & 9.60 & \textbf{3.20} & \textbf{1.49} \\
    \texttt{hex}        & 5.37 & 127.4 & 28.3 & 69.9 & 2.64 & 82.33 & 3.75 & 3.56 & 32.0 & 6.14 &\textbf{ 2.29} & \textbf{0.79} \\
    \texttt{cylinder}   & 4.77 & 130.7 & 21.8 & 146.7 & 4.43 & 155.9 & 4.86 & 8.98 & 8.54 & \textbf{0.03} & \textbf{3.89} & 1.01 \\
    \texttt{semisphere} & 39.7 & 72.0 & 41.6 & 118.9 & 41.8 & 56.5 & 12.3 & 4.35 & 32.4 & 1.51 & \textbf{9.64} & \textbf{0.66} \\
    \texttt{average}    & 46.3 & 105.4 & 52.5 & 111.6 & 43.0 & 78.6 & 7.13 & 4.77 & 23.8 & 8.84 & \textbf{4.04} & \textbf{0.91} \\
    \bottomrule
  \end{tabular}
  \caption{\textbf{Pose Estimation Results.} Translational and rotational errors are shown for the ICP baseline and ViTaSCOPE.}
  \label{tab:ablation_pose_est}
\end{table*}

We design the experiments to test ViTaSCOPE’s ability to:
\begin{itemize}[leftmargin=*]
    \item Accurately reconstruct 3D object geometry using implicit representations.
    \item Estimate in-hand object pose from partial visuo-tactile point cloud observations.
    \item Localize extrinsic contact patches by reasoning over in-hand object poses and distributed tactile shear fields.
\end{itemize}
We evaluate in both simulation (on a held-out test set) and real-world environments. The tools we evaluate over are shown in Fig.~\ref{fig:tool_shapes} placed in the ascending order of complexity in estimating pose and contact. We identify complexity through the number of possible points, lines or small patch contacts. 

\subsection{Baselines}
\begin{itemize}[leftmargin=*]
    \item \textbf{ICP.} To compare the performance of ViTaSCOPE on pose estimation, we baseline our method against a model-based method (ICP~\cite{icp}) implemented by \texttt{open3d}.
    \item \textbf{Neural Contact Fields.} For the extrinsic contact prediction task, we set the SOTA baseline algorithm Neural Contact Fields (NCF) [16] as the baseline, reported in Tab.~\ref{tab:abl} (\textit{ncf baseline}). NCF assumes rigid grasps and known, fixed in-hand object poses during training and testing. It also relies on synthetic point clouds sampled from object meshes at known poses, rather than real visual input and has only been evaluated in sim [16]. However, ViTaSCOPE eliminates these assumptions by handling unknown object poses and fusing real-world vision and touch. Therefore, we run our pose estimation to provide NCF with the object pose to sample a point cloud from the object mesh model as their input as we do not assume fixed in-hand pose. We trained the full NCF model for each geometry by first training the NDF occupancy field, fine-tuning the tactile RGB autoencoder, and the sequence-to-sequence tactile autoencoder.
\end{itemize}

\subsection{Results}
\noindent
\textbf{Recontruction:} To evaluate reconstruction fidelity, we measure the Chamfer Distance (CD) between the normalized SDF mesh reconstruction and the ground truth normalized object mesh by sampling points from the surface as shown in Tab.~1. Here, the normalized geometry means scaling to fit into a unit sphere (r = 1m) as discussed in Section \ref{sec:data_gen}. The mesh reconstructions at the resolution of $256^3$ is shown in Fig.~\ref{fig:tool_shapes}. Chamfer Distance is a commonly used relative metric for evaluating how similar two 3D point clouds (or surfaces) are. This metric measures how close each point in one set is to the nearest point in the other set and averages this over all points. The mean CD of 0.052 $\text{m}^2$ from Tab.~\ref{tab:main_results} suggest that model is able to reconstruct the geometry of all grasped objects accurately.

\vspace{2mm} \noindent
\textbf{In-hand Pose Estimation:} We evaluate the in-hand object pose estimates using Mean Squared Error (MSE) in translation (mm) and angular distance (degrees) compared to ground truth object pose in Tab.~\ref{tab:ablation_pose_est}. The results suggest that our model is able to accurately localize the object despite its partial view of the object. The insight is that tactile sensing provides key contextual cues that can significantly improve pose estimation. The ablation study in Tab. \ref{tab:ablation_pose_est} points out how vision can drive the global alignment of a pose estimate while touch provides dense local points with lower uncertainty as the points lie on the intrinsic contact surface. It can be seen how tactile-only pose estimation can be inherently ambiguous while complementary visuo-tactile pose estimation can refine the pose to improve the accuracy even further. To provide context, the average object is 90 mm, thus the error is in the order of a 2 to 4\% of the object length. The ground truth pose in the real-world is collected using AprilTags which themselves have approximately 1 mm error in our setup. It is also important to note the relative similarity in estimates across simulation and the real-world, suggesting the models ability to effectively bridge the sim-to-real gap as it operates on point clouds. In the open3d implementation of the ICP baseline, the algorithm takes in a segmented point cloud of the object mesh (SDF reconstructed mesh in this case) and the ground truth visuo-tactile point cloud. We emphasize that the comparison is not exactly fair as our model does not have access to the ground truth geometry, instead it approximates it. Despite this, our model outperforms ICP as shown in Tab. \ref{tab:ablation_pose_est}. This table shows how vision or touch alone can introduce ambiguity in object pose whereas when combined the performance becomes more robust. 

\begin{figure*}[t!]
    \centering
    \includegraphics[width=\linewidth]{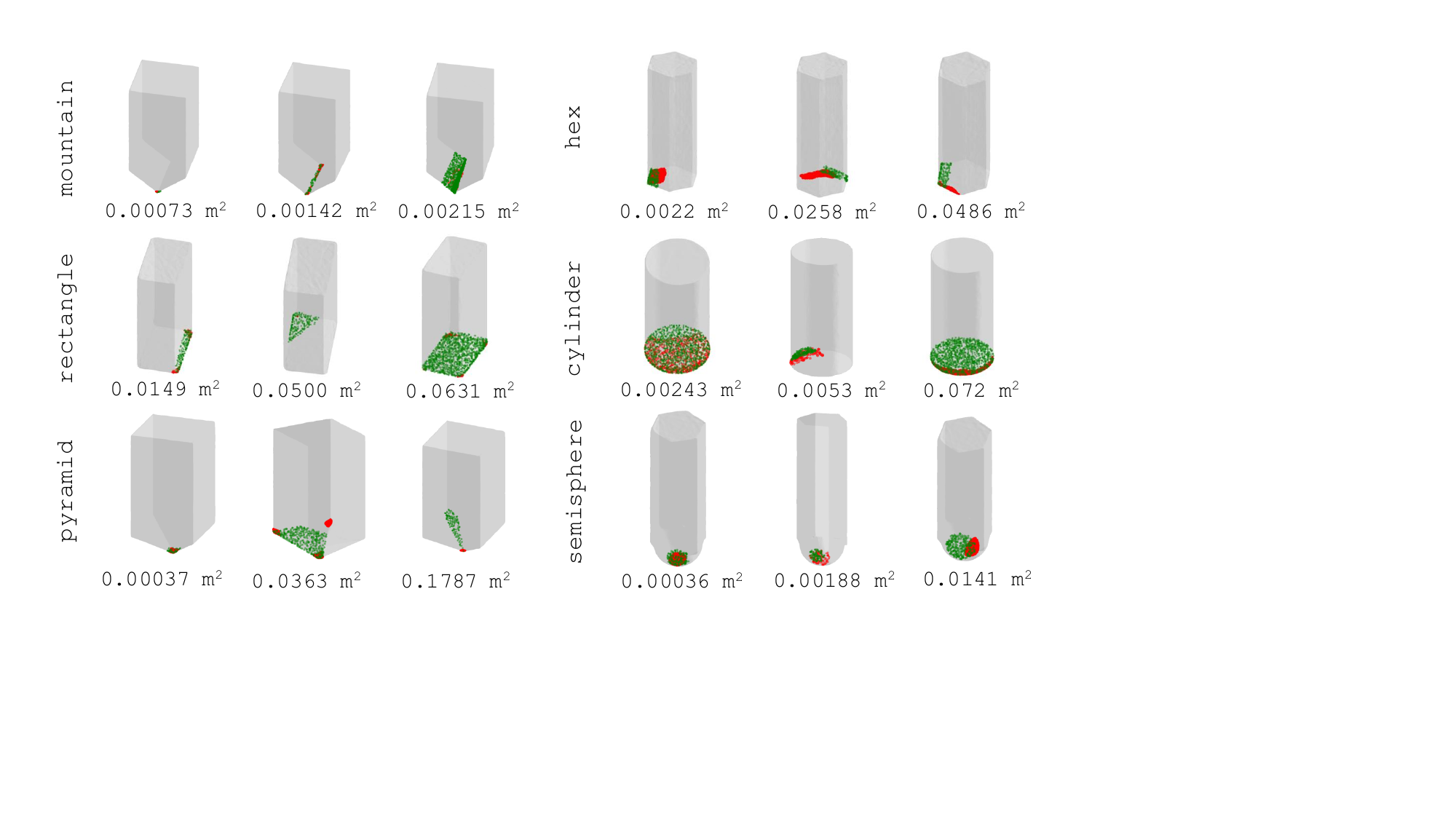}
    \caption{\textbf{Extrinsic Contact Estimation.} Results with varying Chamfer distances and different contact configurations for six tool geometries on the real world test data are shown. The \textcolor{red}{predicted contact patches} are represented with the red pointclouds and the \textcolor{lightgreen}{ground truth contact patches} are in green. The values below each mesh are the CD between the predicted and ground truth contact patches on normalized geometries. ViTaSCOPE's contact predictions are more realistic than the ground truth contact patches obtained via soft contact SDF thresholding (i.e. most rigid-body contacts are point or line). Geometries such as the rectangle and cylinder allows for large contact patches, which tends to result in a higher CD.}
    \label{fig:contact_patches}
\end{figure*}

\vspace{2mm} \noindent
\textbf{Extrinsic Contact Estimation:} We evaluate the extrinsic contact patch using the Chamfer Distance, which is based on normalized geometries. Extrinsic contact estimation is particularly challenging as it often is heavily occluded by the object and potentially the robot itself. Extrinsic contact is also fundamentally ambiguous given purely tactile or force feedback, particularly when vision is deprived. Despite this, ViTaSCOPE accurately predicts extrinsic contact patches, even under challenging occlusions. The contact localization error remains low, with Chamfer Distances comparable between simulation and real-world trials. We note that Isaac Gym does not provide contact locations or wrench and we use mesh and its poses to extract samples of contact points, which uses soft contact thresholding using SDF. Our model learns a better approximation to the contact patch that is closer to realistic contacts than the ground truth soft contact labels as shown in Fig. \ref{fig:contact_patches}. The observed discrepancies in CD arise from soft-contact thresholding between two rigid-body SDFs for generating contact patch labels. However, this results in exaggerated CD values for rectangles and cylinders, which have large ground truth contact patches on large flat surfaces despite ViTaSCOPE's more realistic predictions aligned with rigid-body contacts. With the NCF baseline, we noticed a significant drop in performance during sim-to-real transfer. We attribute this to its sensitivity to pose estimation errors and limited ability to interpret tactile feedback that varies with in-hand pose along with substantial sim-to-real gap in tactile RGB images. In the following Section \ref{sec:abl}, we discuss how our experiments suggest that the tactile shear representations can be more effective for extrinsic contact detection and sim-to-real transfer.

\begin{figure}[b]
    \centering
    \includegraphics[height=2.8cm]{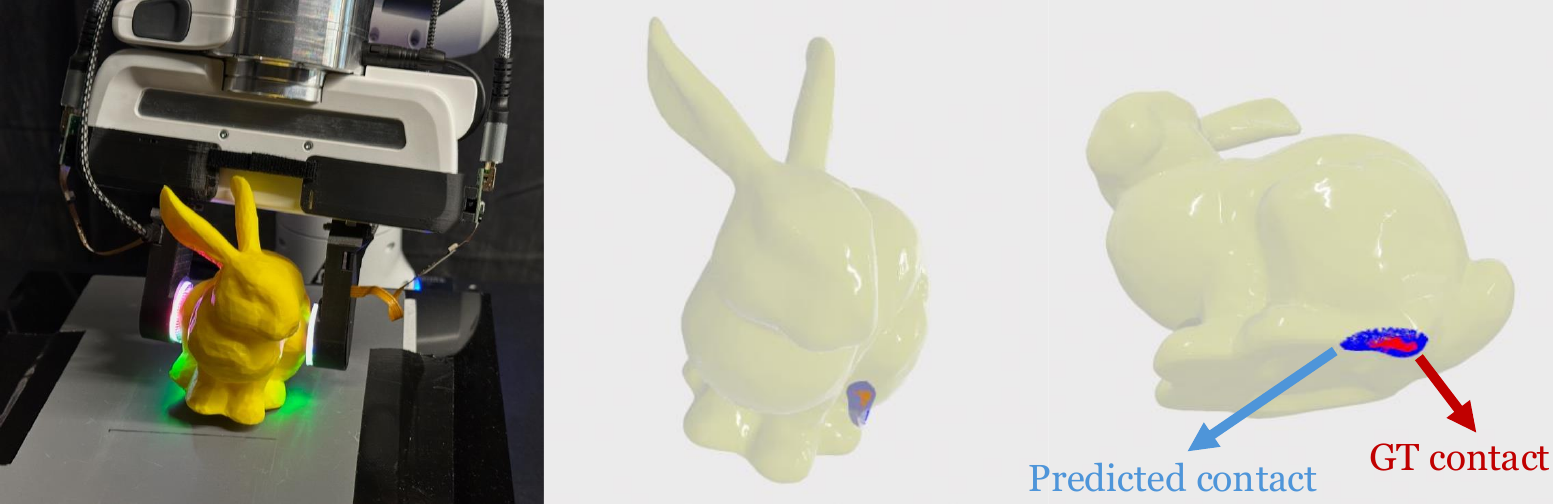}
    \caption{\small\textbf{Sample real-world contact prediction on Stanford bunny.}}
    \label{fig:bunny}
\end{figure}

\begin{table}[t]
  \centering
  \vspace{0.15cm}
  \setlength{\tabcolsep}{2pt}  
  \begin{tabular}{lccccccc}
    \toprule
    Ablation & \multicolumn{7}{c}{Real World Contact Patch Chamfer Distance (CD) $\downarrow$ [$\text{m}^2$]} \\
    \cmidrule(lr){2-8}
             & mount. & rectangle & pyramid & hex & cylinder & semisphere & mean \\
    \midrule
    vitascope        & \textbf{0.0394}  & \textbf{0.2216} & \textbf{0.0565}  & \textbf{0.0596}  & 0.1716 & \textbf{0.0031} & \textbf{0.0920} \\
    wo acts          & 0.0872  & 0.3274 & 0.0992  & 0.0953  & 0.2033 & 0.0044   & 0.1361 \\
    wo obj pose      & 0.1650 & 0.2900 & 0.1335 & 0.1711 & 0.1101 & 0.0073 & 0.1462 \\
    shear $\rightarrow$ rgb       & 0.0551  & 0.2891 & 0.1175 & 0.0611  & \textbf{0.0511}  & 0.0034   & 0.0962 \\
    V pcd & 0.0397  & 0.2227 & 0.0570 & 0.0654  & 0.2580  & 0.0074   & 0.1084 \\
    T pcd & 0.1386 & 0.3720 & 0.0889 & 0.0735  & 0.2145  & 0.0031   & 0.1484 \\
    ncf baseline     & 1.2068  & 0.2452  & 0.6300  & 1.1434  &  0.5930  & 1.5240 &  0.8904 \\
    \bottomrule
  \end{tabular}
  \caption{\small \textbf{Ablation and baseline results for real-world extrinsic contact estimation.} We report the performance measured by CD between predicted and GT contact patches.}
  \label{tab:abl}
  \vspace{-0.6cm}
\end{table}

\subsection{Ablations}
\label{sec:abl}
To understand the contribution of each component in ViTaSCOPE, we conduct a series of ablation studies  and provide Tab.~\ref{tab:abl} as a summary. Overall, our main \textit{vitascope} outperforms several ablated variants of ViTaSCOPE. 

\vspace{2mm} \noindent
\textbf{Effect of Joint Modeling Object Geometry and Pose:} We examine whether the joint modeling of object geometry and pose improves downstream contact prediction by evaluating ViTaSCOPE trained without intermediate activations from the object module (denoted as \textit{wo acts}) and without object pose (denoted as \textit{wo obj pose}). The results support our claim that coarse-to-fine geometrical understanding through the activations and being aware of the in-hand pose is crucial for inferring contact patches from visuo-tactile feedback.

\begin{table}[b]
  \centering
  \small
  \begin{tabular}{lccc}
    \toprule
    Stage & Pose Est. & Trial Code Inf. & Contact Pred. \\
    \midrule
    Time / Steps & 2.5 / 250 & 0.7 / 60 & 0.08 / 1 \\
    \bottomrule
  \end{tabular}
  \caption{\small \textbf{Mean computation time [s] per inference stage.}}
  \label{tab:inf_time}
\end{table}


\vspace{2mm} \noindent
\textbf{Effect of Tactile Shear Representation:} We assess whether the tactile shear representation is suitable for capturing diverse contact modes---particularly in the context of sim-to-real transfer---by replacing the shear field tactile module with an RGB-based one (\textit{shear $\rightarrow$ rgb}). Training this model using tactile RGB images from our simulation dataset results in a performance drop despite using example-based (i.e. uses real sensor's background image) realistic RGB tactile simulation [1,11]. This is likely due to its inability to capture intricate and subtle extrinsic contact cues compared to shear representation, which enables cross-sensory generalization (GelSight in sim to GelSlim in real) and sim-to-real transfer.

\vspace{2mm} \noindent
\textbf{Effect of In-Hand Pose Accuracy on Contact Prediction:} In Tab. \ref{tab:ablation_pose_est} shows that the pose accuracy degrades when using vision-only (denoted as \textit{V pcd}) or tactile-only (denoted as \textit{T pcd}) point clouds. We extend this analysis to assess the impact on downstream extrinsic contact prediction (Tab.~\ref{tab:abl}). Running inference on vision-only pose estimate (\textit{V pcd}), despite being slightly less accurate, performs comparably to the visuo-tactile \textit{vitascope} model, demonstrating ViTaSCOPE's robustness to moderate pose errors. In contrast, running inference with tactile-only pose estimate (\textit{T pcd}) yields performance similar to the model without object pose (\textit{wo obj pose}), underscoring the critical role of complementary visuo-tactile feedback for accurate pose estimation and effective contact prediction.

\vspace{2mm} \noindent
\textbf{Inference Time and Generalizability:} We outline the computational time of our two-stage inference (Pose Est. \& Trial Code Inf.) and the number of gradient steps per optimization and forward pass (Contact Pred.) in Tab.~\ref{tab:inf_time}. Our model runs at 0.3Hz, with faster inference possible by reducing optimization steps or tuning the optimizer, at some cost to accuracy. To demonstrate ViTaSCOPE's ability to handle more complex geometries, we generated simulation data and trained it on Stanford bunny and show a real-world prediction in Fig.~\ref{fig:bunny}. For generalization to unseen objects, we can use object codes via hyper-networks trained on large-scale object sets, as in DeepSDF [14]. Open-vocabulary detectors can jumpstart object code inference.
\section{Discussion and Limitations}
\vspace{2mm} \noindent
\textbf{Summary:}
ViTaSCOPE is a unified framework for in-hand pose estimation and extrinsic contact localization, leveraging implicit representations to seamlessly integrate visuo-tactile feedback. Our results demonstrate that tactile shear fields offer rich information for estimating contact distributions and, when combined with visual input, can provide robust geometric reasoning for manipulation tasks. Notably, our approach generalizes across different tool geometries and contact interactions.

As expected, we observe experimentally that vision and touch provide complementary information that enhances both global spatial reasoning and local contact understanding. Vision provides a broad, albeit often occluded, view of the scene, while tactile sensing offers high-resolution measurements of local shear forces induced by extrinsic contacts. By integrating these signals within an implicit function framework, ViTaSCOPE bridges the gap between global object geometry and fine-grained contact details, enabling simultaneous state estimation and interaction modeling.

Additionally, our sim-to-real approach demonstrates the feasibility of training contact-aware models in simulation while preserving their effectiveness in real-world manipulation tasks. 

\vspace{2mm} \noindent
\textbf{Limitations:}
\textbf{Assumption of High-Fidelity Tactile Data.} Our method relies on accurate shear field measurements from vision-based tactile sensors. However, sensor noise, lighting conditions, and surface compliance variations can introduce artifacts, affecting estimation accuracy.

\vspace{2mm} \noindent
\textbf{Lack of Normal Force Information.} We currently focus on shear fields and do not explicitly model normal forces, which are crucial for capturing contact pressures and forces perpendicular to the surface. Future extensions could incorporate normal force estimation to improve contact modeling.

\vspace{2mm} \noindent
\textbf{Handling of Complex Contact Interactions.} While ViTaSCOPE can generalize across different geometries, certain ambiguous contact configurations (e.g., multiple small contact patches producing similar shear responses) remain difficult to disambiguate. This ambiguity is likely fundamental and unlikely to be overcome with conventional tactile sensing technologies.

\vspace{2mm} \noindent
\textbf{In-Hand Slippage and Deformation.} Gel-based tactile sensors introduce compliance, meaning object pose may shift subtly during contact interactions. Here, we do not address the possibility of slippage between the robot and object.

\vspace{2mm} \noindent
\textbf{Scaling to More General Objects.} Our framework assumes that object geometries are known at training time. Generalizing to previously unseen objects without requiring explicit mesh models is an important direction for future work.

\vspace{2mm} \noindent
\textbf{Extending to Dexterous Hands.} Extending ViTaSCOPE to a dexterous multi-fingered hand with our distributed tactile module would be an interesting direction to take. 

\section*{Acknowledgments}
We would like to thank Andrea Sipos for manufacturing the Gelslim 4.0 vision-based tactile sensors for the Franka Emika Panda robot arm. We also thank William van den Bogert for his help with tracking marker displacements from Gelslim 4.0 to extract shear observations. We are grateful to Mark van der Merwe, Youngsun Wi, Miquel Oller, and Samanta Rodriguez for helpful discussions and important feedbacks. \\

\bibliographystyle{unsrtnat}
\bibliography{references}

\begin{thebibliography}{46}
\providecommand{\natexlab}[1]{#1}
\providecommand{\url}[1]{\texttt{#1}}
\expandafter\ifx\csname urlstyle\endcsname\relax
  \providecommand{\doi}[1]{doi: #1}\else
  \providecommand{\doi}{doi: \begingroup \urlstyle{rm}\Url}\fi

\bibitem[Akinola et~al.(2024)Akinola, Xu, Carius, Fox, and Narang]{tacsl}
Iretiayo Akinola, Jie Xu, Jan Carius, Dieter Fox, and Yashraj~S. Narang.
\newblock \href{https://doi.org/10.48550/arXiv.2408.06506}{TacSL: {A} Library for Visuotactile Sensor Simulation and Learning}.
\newblock \emph{CoRR}, abs/2408.06506, 2024.

\bibitem[Lambeta et~al.(2020)Lambeta, Chou, Tian, Yang, Maloon, Most, Stroud, Santos, Byagowi, Kammerer, Jayaraman, and Calandra]{digit}
Mike Lambeta, Po{-}Wei Chou, Stephen Tian, Brian~H. Yang, Benjamin Maloon, Victoria~Rose Most, Dave Stroud, Raymond Santos, Ahmad Byagowi, Gregg Kammerer, Dinesh Jayaraman, and Roberto Calandra.
\newblock \href{https://doi.org/10.1109/LRA.2020.2977257}{{DIGIT:} {A} Novel Design for a Low-Cost Compact High-Resolution Tactile Sensor With Application to In-Hand Manipulation}.
\newblock \emph{{IEEE} Robotics Autom. Lett.}, 5\penalty0 (3):\penalty0 3838--3845, 2020.
\newblock \doi{10.1109/LRA.2020.2977257}.

\bibitem[Li et~al.(2014)Li, Jr., Yuan, ten Pas, Roscup, Srinivasan, and Adelson]{gelsight}
Rui Li, Robert~Platt Jr., Wenzhen Yuan, Andreas ten Pas, Nathan Roscup, Mandayam~A. Srinivasan, and Edward~H. Adelson.
\newblock \href{https://doi.org/10.1109/IROS.2014.6943123}{Localization and manipulation of small parts using GelSight tactile sensing}.
\newblock In \emph{2014 {IEEE/RSJ} International Conference on Intelligent Robots and Systems, {IROS} 2014, Chicago, IL, USA, September 14-18, 2014}, pages 3988--3993. {IEEE}, 2014.
\newblock \doi{10.1109/IROS.2014.6943123}.

\bibitem[Taylor et~al.(2022)Taylor, Dong, and Rodriguez]{gelslim3}
Ian~H. Taylor, Siyuan Dong, and Alberto Rodriguez.
\newblock \href{https://doi.org/10.1109/ICRA46639.2022.9811832}{GelSlim 3.0: High-Resolution Measurement of Shape, Force and Slip in a Compact Tactile-Sensing Finger}.
\newblock In \emph{2022 International Conference on Robotics and Automation, {ICRA} 2022, Philadelphia, PA, USA, May 23-27, 2022}, pages 10781--10787. {IEEE}, 2022.
\newblock \doi{10.1109/ICRA46639.2022.9811832}.

\bibitem[Sipos et~al.(2024)Sipos, van~den Bogert, and Fazeli]{gelslim4}
Andrea Sipos, William van~den Bogert, and Nima Fazeli.
\newblock \href{https://doi.org/10.48550/arXiv.2409.19770}{GelSlim 4.0: Focusing on Touch and Reproducibility}.
\newblock \emph{CoRR}, abs/2409.19770, 2024.
\newblock \doi{10.48550/ARXIV.2409.19770}.

\bibitem[Alspach et~al.(2019)Alspach, Hashimoto, Kuppuswamy, and Tedrake]{softbubble}
Alex Alspach, Kunimatsu Hashimoto, Naveen Kuppuswamy, and Russ Tedrake.
\newblock \href{https://doi.org/10.1109/ROBOSOFT.2019.872271}{Soft-bubble: {A} highly compliant dense geometry tactile sensor for robot manipulation}.
\newblock In \emph{{IEEE} International Conference on Soft Robotics, RoboSoft 2019, Seoul, South Korea, April 14-18, 2019}, pages 597--604. {IEEE}, 2019.
\newblock \doi{10.1109/ROBOSOFT.2019.8722713}.

\bibitem[Lambeta et~al.(2024)Lambeta, Wu, Sengul, Most, Black, Sawyer, Mercado, Qi, Sohn, Taylor, Tydingco, Kammerer, Stroud, Khatha, Jenkins, Most, Stein, Chavira, Craven{-}Bartle, Sanchez, Ding, Malik, and Calandra]{digit360}
Mike Lambeta, Tingfan Wu, Ali Sengul, Victoria~Rose Most, Nolan Black, Kevin Sawyer, Romeo Mercado, Haozhi Qi, Alexander Sohn, Byron Taylor, Norb Tydingco, Gregg Kammerer, Dave Stroud, Jake Khatha, Kurt Jenkins, Kyle Most, Neal Stein, Ricardo Chavira, Thomas Craven{-}Bartle, Eric Sanchez, Yitian Ding, Jitendra Malik, and Roberto Calandra.
\newblock \href{https://doi.org/10.48550/arXiv.2411.02479}{Digitizing Touch with an Artificial Multimodal Fingertip}.
\newblock \emph{CoRR}, abs/2411.02479, 2024.
\newblock \doi{10.48550/ARXIV.2411.02479}.

\bibitem[Wang et~al.(2022)Wang, Lambeta, Chou, and Calandra]{tacto}
Shaoxiong Wang, Mike Lambeta, Po{-}Wei Chou, and Roberto Calandra.
\newblock \href{https://doi.org/10.1109/LRA.2022.3146945}{{TACTO:} {A} Fast, Flexible, and Open-Source Simulator for High-Resolution Vision-Based Tactile Sensors}.
\newblock \emph{{IEEE} Robotics Autom. Lett.}, 7\penalty0 (2):\penalty0 3930--3937, 2022.
\newblock \doi{10.1109/LRA.2022.3146945}.

\bibitem[Xu et~al.(2022)Xu, Kim, Chen, Garcia, Agrawal, Matusik, and Sueda]{tactilesim}
Jie Xu, Sangwoon Kim, Tao Chen, Alberto~Rodriguez Garcia, Pulkit Agrawal, Wojciech Matusik, and Shinjiro Sueda.
\newblock \href{https://proceedings.mlr.press/v205/xu23b.html}{Efficient Tactile Simulation with Differentiability for Robotic Manipulation}.
\newblock In \emph{Conference on Robot Learning, CoRL 2022, 14-18 December 2022, Auckland, New Zealand}, volume 205 of \emph{Proceedings of Machine Learning Research}, pages 1488--1498. {PMLR}, 2022.

\bibitem[Makoviychuk et~al.(2021)Makoviychuk, Wawrzyniak, Guo, Lu, Storey, Macklin, Hoeller, Rudin, Allshire, Handa, and State]{isaacgym}
Viktor Makoviychuk, Lukasz Wawrzyniak, Yunrong Guo, Michelle Lu, Kier Storey, Miles Macklin, David Hoeller, Nikita Rudin, Arthur Allshire, Ankur Handa, and Gavriel State.
\newblock \href{https://datasets-benchmarks-proceedings.neurips.cc/paper/2021/hash/28dd2c7955ce926456240b2ff0100bde-Abstract-round2.html}{Isaac Gym: High Performance {GPU} Based Physics Simulation For Robot Learning}.
\newblock In \emph{Proceedings of the Neural Information Processing Systems Track on Datasets and Benchmarks 1, NeurIPS Datasets and Benchmarks 2021, December 2021, virtual}, 2021.

\bibitem[Si and Yuan(2022)]{taxim}
Zilin Si and Wenzhen Yuan.
\newblock \href{https://doi.org/10.1109/LRA.2022.3142412}{Taxim: An Example-Based Simulation Model for GelSight Tactile Sensors}.
\newblock \emph{{IEEE} Robotics Autom. Lett.}, 7\penalty0 (2):\penalty0 2361--2368, 2022.
\newblock \doi{10.1109/LRA.2022.3142412}.

\bibitem[Si et~al.(2024)Si, Zhang, Ben, Romero, Liu, and Gan]{difftactile}
Zilin Si, Gu~Zhang, Qingwei Ben, Branden Romero, Chao Liu, and Chuang Gan.
\newblock \href{https://openreview.net/forum?id=eJHnSg783t}{{DIFFTACTILE:} {A} Physics-based Differentiable Tactile Simulator for Contact-rich Robotic Manipulation}.
\newblock In \emph{The Twelfth International Conference on Learning Representations, {ICLR} 2024, Vienna, Austria, May 7-11, 2024}. OpenReview.net, 2024.

\bibitem[Sferrazza et~al.(2024)Sferrazza, Seo, Liu, Lee, and Abbeel]{m3l}
Carmelo Sferrazza, Younggyo Seo, Hao Liu, Youngwoon Lee, and Pieter Abbeel.
\newblock \href{https://doi.org/10.1109/IROS58592.2024.10802719}{The Power of the Senses: Generalizable Manipulation from Vision and Touch through Masked Multimodal Learning}.
\newblock In \emph{{IEEE/RSJ} International Conference on Intelligent Robots and Systems, {IROS} 2024, Abu Dhabi, United Arab Emirates, October 14-18, 2024}, pages 9698--9705. {IEEE}, 2024.
\newblock \doi{10.1109/IROS58592.2024.10802719}.

\bibitem[Todorov et~al.(2012)Todorov, Erez, and Tassa]{mujoco}
Emanuel Todorov, Tom Erez, and Yuval Tassa.
\newblock \href{https://ieeexplore.ieee.org/document/6386109}{MuJoCo: A physics engine for model-based control}.
\newblock In \emph{2012 IEEE/RSJ International Conference on Intelligent Robots and Systems}, pages 5026--5033. IEEE, 2012.
\newblock \doi{10.1109/IROS.2012.6386109}.

\bibitem[Park et~al.(2019)Park, Florence, Straub, Newcombe, and Lovegrove]{deepsdf}
Jeong~Joon Park, Peter~R. Florence, Julian Straub, Richard~A. Newcombe, and Steven Lovegrove.
\newblock \href{http://openaccess.thecvf.com/content_CVPR_2019/html/Park_DeepSDF_Learning_Continuous_Signed_Distance_Functions_for_Shape_Representation_CVPR_2019_paper.html}{DeepSDF: Learning Continuous Signed Distance Functions for Shape Representation}.
\newblock In \emph{{IEEE} Conference on Computer Vision and Pattern Recognition, {CVPR} 2019, Long Beach, CA, USA, June 16-20, 2019}, pages 165--174. Computer Vision Foundation / {IEEE}, 2019.
\newblock \doi{10.1109/CVPR.2019.00025}.

\bibitem[Wi et~al.(2022{\natexlab{a}})Wi, Florence, Zeng, and Fazeli]{virdo}
Youngsun Wi, Pete Florence, Andy Zeng, and Nima Fazeli.
\newblock \href{https://doi.org/10.1109/ICRA46639.2022.9812097}{VIRDO: Visio-tactile Implicit Representations of Deformable Objects}.
\newblock In \emph{2022 International Conference on Robotics and Automation, {ICRA} 2022, Philadelphia, PA, USA, May 23-27, 2022}, pages 3583--3590. {IEEE}, 2022{\natexlab{a}}.
\newblock \doi{10.1109/ICRA46639.2022.9812097}.

\bibitem[Wi et~al.(2022{\natexlab{b}})Wi, Zeng, Florence, and Fazeli]{virdopp}
Youngsun Wi, Andy Zeng, Pete Florence, and Nima Fazeli.
\newblock \href{https://proceedings.mlr.press/v205/wi23a.html}{VIRDO++: Real-World, Visuo-tactile Dynamics and Perception of Deformable Objects}.
\newblock In \emph{Conference on Robot Learning, CoRL 2022, 14-18 December 2022, Auckland, New Zealand}, volume 205 of \emph{Proceedings of Machine Learning Research}, pages 1806--1816. {PMLR}, 2022{\natexlab{b}}.

\bibitem[der Merwe et~al.(2023)der Merwe, Wi, Berenson, and Fazeli]{ndcf}
Mark J.~Van der Merwe, Youngsun Wi, Dmitry Berenson, and Nima Fazeli.
\newblock \href{https://doi.org/10.15607/RSS.2023.XIX.080}{Integrated Object Deformation and Contact Patch Estimation from Visuo-Tactile Feedback}.
\newblock In \emph{Robotics: Science and Systems XIX, Daegu, Republic of Korea, July 10-14, 2023}, 2023.
\newblock \doi{10.15607/RSS.2023.XIX.080}.

\bibitem[Higuera et~al.(2023{\natexlab{a}})Higuera, Dong, Boots, and Mukadam]{ncf}
Carolina Higuera, Siyuan Dong, Byron Boots, and Mustafa Mukadam.
\newblock \href{https://doi.org/10.1109/ICRA48891.2023.10160526}{Neural Contact Fields: Tracking Extrinsic Contact with Tactile Sensing}.
\newblock In \emph{{IEEE} International Conference on Robotics and Automation, {ICRA} 2023, London, UK, May 29 - June 2, 2023}, pages 12576--12582. {IEEE}, 2023{\natexlab{a}}.
\newblock \doi{10.1109/ICRA48891.2023.10160526}.

\bibitem[Wi et~al.(2025)Wi, Lee, Oller, and Fazeli]{nisp}
Youngsun Wi, Jayjun Lee, Miquel Oller, and Nima Fazeli.
\newblock \href{https://proceedings.mlr.press/v270/wi25a.html}{Neural Inverse Source Problem}.
\newblock In \emph{Proceedings of The 8th Conference on Robot Learning}, volume 270 of \emph{Proceedings of Machine Learning Research}, pages 4371--4391. PMLR, 06--09 Nov 2025.

\bibitem[Chen et~al.(2022)Chen, Hasson, Schmid, and Laptev]{alignsdf}
Zerui Chen, Yana Hasson, Cordelia Schmid, and Ivan Laptev.
\newblock \href{https://doi.org/10.1007/978-3-031-19769-7_14}{AlignSDF: Pose-Aligned Signed Distance Fields for Hand-Object Reconstruction}.
\newblock In \emph{Computer Vision - {ECCV} 2022 - 17th European Conference, Tel Aviv, Israel, October 23-27, 2022, Proceedings, Part {I}}, volume 13661 of \emph{Lecture Notes in Computer Science}, pages 231--248. Springer, 2022.
\newblock \doi{10.1007/978-3-031-19769-7\_14}.

\bibitem[Wen et~al.(2024)Wen, Yang, Kautz, and Birchfield]{foundationpose}
Bowen Wen, Wei Yang, Jan Kautz, and Stan Birchfield.
\newblock \href{https://doi.org/10.1109/CVPR52733.2024.01692}{FoundationPose: Unified 6D Pose Estimation and Tracking of Novel Objects}.
\newblock In \emph{{IEEE/CVF} Conference on Computer Vision and Pattern Recognition, {CVPR} 2024, Seattle, WA, USA, June 16-22, 2024}, pages 17868--17879. {IEEE}, 2024.
\newblock \doi{10.1109/CVPR52733.2024.01692}.

\bibitem[Ye et~al.(2022)Ye, Gupta, and Tulsiani]{whatsinyourhand}
Yufei Ye, Abhinav Gupta, and Shubham Tulsiani.
\newblock \href{https://doi.org/10.1109/CVPR52688.2022.00387}{What's in your hands? 3D Reconstruction of Generic Objects in Hands}.
\newblock In \emph{{IEEE/CVF} Conference on Computer Vision and Pattern Recognition, {CVPR} 2022, New Orleans, LA, USA, June 18-24, 2022}, pages 3885--3895. {IEEE}, 2022.
\newblock \doi{10.1109/CVPR52688.2022.00387}.

\bibitem[Bauz{\'{a}} et~al.(2023)Bauz{\'{a}}, Bronars, and Rodriguez]{tac2pose}
Maria Bauz{\'{a}}, Antonia Bronars, and Alberto Rodriguez.
\newblock \href{https://doi.org/10.1177/02783649231196925}{Tac2Pose: Tactile object pose estimation from the first touch}.
\newblock \emph{Int. J. Robotics Res.}, 42\penalty0 (13):\penalty0 1185--1209, 2023.
\newblock \doi{10.1177/02783649231196925}.

\bibitem[Kuppuswamy et~al.(2020)Kuppuswamy, Castro, Phillips{-}Grafflin, Alspach, and Tedrake]{bubblecontactpatch}
Naveen Kuppuswamy, Alejandro~M. Castro, Calder Phillips{-}Grafflin, Alex Alspach, and Russ Tedrake.
\newblock \href{https://doi.org/10.1109/LRA.2019.2961050}{Fast Model-Based Contact Patch and Pose Estimation for Highly Deformable Dense-Geometry Tactile Sensors}.
\newblock \emph{{IEEE} Robotics Autom. Lett.}, 5\penalty0 (2):\penalty0 1811--1818, 2020.
\newblock \doi{10.1109/LRA.2019.2961050}.

\bibitem[Izatt et~al.(2017)Izatt, Mirano, Adelson, and Tedrake]{trackingobjectsruss}
Gregory Izatt, Geronimo Mirano, Edward~H. Adelson, and Russ Tedrake.
\newblock \href{https://doi.org/10.1109/ICRA.2017.7989460}{Tracking objects with point clouds from vision and touch}.
\newblock In \emph{2017 {IEEE} International Conference on Robotics and Automation, {ICRA} 2017, Singapore, Singapore, May 29 - June 3, 2017}, pages 4000--4007. {IEEE}, 2017.
\newblock \doi{10.1109/ICRA.2017.7989460}.

\bibitem[Smith et~al.(2020)Smith, Calandra, Romero, Gkioxari, Meger, Malik, and Drozdzal]{shaperecon}
Edward~J. Smith, Roberto Calandra, Adriana Romero, Georgia Gkioxari, David Meger, Jitendra Malik, and Michal Drozdzal.
\newblock \href{https://proceedings.neurips.cc/paper/2020/hash/a3842ed7b3d0fe3ac263bcabd2999790-Abstract.html}{3D Shape Reconstruction from Vision and Touch}.
\newblock In \emph{Advances in Neural Information Processing Systems 33: Annual Conference on Neural Information Processing Systems 2020, NeurIPS 2020, December 6-12, 2020, virtual}, 2020.

\bibitem[Smith et~al.(2021)Smith, Meger, Calandra, Malik, Romero{-}Soriano, and Drozdzal]{activeshaperecon}
Edward~J. Smith, David Meger, Luis Pineda~Roberto Calandra, Jitendra Malik, Adriana Romero{-}Soriano, and Michal Drozdzal.
\newblock \href{https://proceedings.neurips.cc/paper/2021/hash/8635b5fd6bc675033fb72e8a3ccc10a0-Abstract.html}{Active 3D Shape Reconstruction from Vision and Touch}.
\newblock In \emph{Advances in Neural Information Processing Systems 34: Annual Conference on Neural Information Processing Systems 2021, NeurIPS 2021, December 6-14, 2021, virtual}, pages 16064--16078, 2021.

\bibitem[Li et~al.(2023)Li, Dikhale, Iba, and Jamali]{vihope}
Hongyu Li, Snehal Dikhale, Soshi Iba, and Nawid Jamali.
\newblock \href{https://doi.org/10.1109/LRA.2023.3313941}{ViHOPE: Visuotactile In-Hand Object 6D Pose Estimation With Shape Completion}.
\newblock \emph{{IEEE} Robotics Autom. Lett.}, 8\penalty0 (11):\penalty0 6963--6970, 2023.
\newblock \doi{10.1109/LRA.2023.3313941}.

\bibitem[Dikhale et~al.(2022)Dikhale, Patel, Dhingra, Naramura, Hayashi, Iba, and Jamali]{visuotactile6dposeest}
Snehal Dikhale, Karankumar Patel, Daksh Dhingra, Itoshi Naramura, Akinobu Hayashi, Soshi Iba, and Nawid Jamali.
\newblock \href{https://doi.org/10.1109/LRA.2022.3143289}{VisuoTactile 6D Pose Estimation of an In-Hand Object Using Vision and Tactile Sensor Data}.
\newblock \emph{{IEEE} Robotics Autom. Lett.}, 7\penalty0 (2):\penalty0 2148--2155, 2022.
\newblock \doi{10.1109/LRA.2022.3143289}.

\bibitem[Sipos and Fazeli(2022)]{scope}
Andrea Sipos and Nima Fazeli.
\newblock \href{https://doi.org/10.1109/IROS47612.2022.9981762}{Simultaneous Contact Location and Object Pose Estimation Using Proprioception and Tactile Feedback}.
\newblock In \emph{{IEEE/RSJ} International Conference on Intelligent Robots and Systems, {IROS} 2022, Kyoto, Japan, October 23-27, 2022}, pages 3233--3240. {IEEE}, 2022.
\newblock \doi{10.1109/IROS47612.2022.9981762}.

\bibitem[Sipos and Fazeli(2023)]{multiscope}
Andrea Sipos and Nima Fazeli.
\newblock \href{https://doi.org/10.15607/RSS.2023.XIX.078}{MultiSCOPE: Disambiguating In-Hand Object Poses with Proprioception and Tactile Feedback}.
\newblock In \emph{Robotics: Science and Systems XIX, Daegu, Republic of Korea, July 10-14, 2023}, 2023.
\newblock \doi{10.15607/RSS.2023.XIX.078}.

\bibitem[Manuelli and Tedrake(2016)]{cpf}
Lucas Manuelli and Russ Tedrake.
\newblock \href{https://doi.org/10.1109/IROS.2016.7759743}{Localizing external contact using proprioceptive sensors: The Contact Particle Filter}.
\newblock In \emph{2016 {IEEE/RSJ} International Conference on Intelligent Robots and Systems, {IROS} 2016, Daejeon, South Korea, October 9-14, 2016}, pages 5062--5069. {IEEE}, 2016.
\newblock \doi{10.1109/IROS.2016.7759743}.

\bibitem[Kim et~al.(2023{\natexlab{a}})Kim, Li, Posa, and Jayaraman]{im2contact}
Leon Kim, Yunshuang Li, Michael Posa, and Dinesh Jayaraman.
\newblock \href{https://proceedings.mlr.press/v229/kim23b.html}{Im2Contact: Vision-Based Contact Localization Without Touch or Force Sensing}.
\newblock In \emph{Conference on Robot Learning, CoRL 2023, 6-9 November 2023, Atlanta, GA, {USA}}, volume 229 of \emph{Proceedings of Machine Learning Research}, pages 1533--1546. {PMLR}, 2023{\natexlab{a}}.

\bibitem[Yi et~al.(2024)Yi, Lee, and Fazeli]{vacontact}
Xili Yi, Jayjun Lee, and Nima Fazeli.
\newblock \href{https://doi.org/10.48550/arXiv.2409.14608}{Visual-auditory Extrinsic Contact Estimation}.
\newblock \emph{CoRR}, abs/2409.14608, 2024.
\newblock \doi{10.48550/ARXIV.2409.14608}.

\bibitem[Higuera et~al.(2023{\natexlab{b}})Higuera, Ortiz, Qi, Pineda, Boots, and Mukadam]{ncfv2}
Carolina Higuera, Joseph Ortiz, Haozhi Qi, Luis Pineda, Byron Boots, and Mustafa Mukadam.
\newblock \href{https://doi.org/10.48550/arXiv.2309.16652}{Perceiving Extrinsic Contacts from Touch Improves Learning Insertion Policies}.
\newblock \emph{CoRR}, abs/2309.16652, 2023{\natexlab{b}}.
\newblock \doi{10.48550/ARXIV.2309.16652}.

\bibitem[Oller et~al.(2022)Oller, Planas, Berenson, and Fazeli]{ollercorl}
Miquel Oller, Mireia Planas, Dmitry Berenson, and Nima Fazeli.
\newblock \href{https://proceedings.mlr.press/v205/oller23a.html}{Manipulation via Membranes: High-Resolution and Highly Deformable Tactile Sensing and Control}.
\newblock In \emph{Conference on Robot Learning, CoRL 2022, 14-18 December 2022, Auckland, New Zealand}, volume 205 of \emph{Proceedings of Machine Learning Research}, pages 1850--1859. {PMLR}, 2022.

\bibitem[Oller et~al.(2024)Oller, Berenson, and Fazeli]{ollerrss}
Miquel Oller, Dmitry Berenson, and Nima Fazeli.
\newblock \href{https://www.roboticsproceedings.org/rss20/p135.pdf}{Tactile-Driven Non-Prehensile Object Manipulation via Extrinsic Contact Mode Control}.
\newblock In \emph{Proceedings of Robotics: Science and Systems}, Delft, Netherlands, July 2024.
\newblock \doi{10.15607/RSS.2024.XX.135}.

\bibitem[Ma et~al.(2021)Ma, Dong, and Rodriguez]{ext_contact_sensing_w_rel_motion}
Daolin Ma, Siyuan Dong, and Alberto Rodriguez.
\newblock \href{https://doi.org/10.1109/ICRA48506.2021.9561781}{Extrinsic Contact Sensing with Relative-Motion Tracking from Distributed Tactile Measurements}.
\newblock In \emph{{IEEE} International Conference on Robotics and Automation, {ICRA} 2021, Xi'an, China, May 30 - June 5, 2021}, pages 11262--11268. {IEEE}, 2021.
\newblock \doi{10.1109/ICRA48506.2021.9561781}.

\bibitem[Dong and Rodriguez(2019)]{tactile_dense_packing}
Siyuan Dong and Alberto Rodriguez.
\newblock \href{https://doi.org/10.1109/IROS40897.2019.8968204}{Tactile-Based Insertion for Dense Box-Packing}.
\newblock In \emph{2019 {IEEE/RSJ} International Conference on Intelligent Robots and Systems, {IROS} 2019, Macau, SAR, China, November 3-8, 2019}, pages 7953--7960. {IEEE}, 2019.
\newblock \doi{10.1109/IROS40897.2019.8968204}.

\bibitem[Ma et~al.(2019)Ma, Donlon, Dong, and Rodriguez]{tactile_force_est_iFEM}
Daolin Ma, Elliott Donlon, Siyuan Dong, and Alberto Rodriguez.
\newblock \href{https://doi.org/10.1109/ICRA.2019.8794113}{Dense Tactile Force Estimation using GelSlim and inverse {FEM}}.
\newblock In \emph{International Conference on Robotics and Automation, {ICRA} 2019, Montreal, QC, Canada, May 20-24, 2019}, pages 5418--5424. {IEEE}, 2019.
\newblock \doi{10.1109/ICRA.2019.8794113}.

\bibitem[Kim et~al.(2023{\natexlab{b}})Kim, Jha, Romeres, Patre, and Rodriguez]{simult_tact_est_and_cont_of_ext_contact}
Sangwoon Kim, Devesh~K. Jha, Diego Romeres, Parag Patre, and Alberto Rodriguez.
\newblock \href{https://doi.org/10.1109/ICRA48891.2023.10161158}{Simultaneous Tactile Estimation and Control of Extrinsic Contact}.
\newblock In \emph{{IEEE} International Conference on Robotics and Automation, {ICRA} 2023, London, UK, May 29 - June 2, 2023}, pages 12563--12569. {IEEE}, 2023{\natexlab{b}}.
\newblock \doi{10.1109/ICRA48891.2023.10161158}.

\bibitem[Bronars et~al.(2024)Bronars, Kim, and Rodriguez]{texterity}
Antonia Bronars, Sangwoon Kim, and Parag Patre~Alberto Rodriguez.
\newblock \href{https://doi.org/10.1109/ICRA57147.2024.10610622}{TEXterity: Tactile Extrinsic deXterity}.
\newblock In \emph{{IEEE} International Conference on Robotics and Automation, {ICRA} 2024, Yokohama, Japan, May 13-17, 2024}, pages 7976--7983. {IEEE}, 2024.
\newblock \doi{10.1109/ICRA57147.2024.10610622}.

\bibitem[Lorensen and Cline(1987)]{marchingcubes}
William~E. Lorensen and Harvey~E. Cline.
\newblock \href{https://doi.org/10.1145/37401.37422}{Marching cubes: {A} high resolution 3D surface construction algorithm}.
\newblock In Maureen~C. Stone, editor, \emph{Proceedings of the 14th Annual Conference on Computer Graphics and Interactive Techniques, {SIGGRAPH} 1987, Anaheim, California, USA, July 27-31, 1987}, pages 163--169. {ACM}, 1987.
\newblock \doi{10.1145/37401.37422}.

\bibitem[Ren et~al.(2024)Ren, Liu, Zeng, Lin, Li, Cao, Chen, Huang, Chen, Yan, Zeng, Zhang, Li, Yang, Li, Jiang, and Zhang]{gsam}
Tianhe Ren, Shilong Liu, Ailing Zeng, Jing Lin, Kunchang Li, He~Cao, Jiayu Chen, Xinyu Huang, Yukang Chen, Feng Yan, Zhaoyang Zeng, Hao Zhang, Feng Li, Jie Yang, Hongyang Li, Qing Jiang, and Lei Zhang.
\newblock \href{https://doi.org/10.48550/arXiv.2401.14159}{Grounded {SAM:} Assembling Open-World Models for Diverse Visual Tasks}.
\newblock \emph{CoRR}, abs/2401.14159, 2024.
\newblock \doi{10.48550/ARXIV.2401.14159}.

\bibitem[Besl and McKay(1992)]{icp}
P.J. Besl and Neil~D. McKay.
\newblock \href{https://ieeexplore.ieee.org/document/121791}{A method for registration of 3-D shapes}.
\newblock \emph{IEEE Transactions on Pattern Analysis and Machine Intelligence}, 14\penalty0 (2):\penalty0 239--256, 1992.
\newblock \doi{10.1109/34.121791}.

\end{thebibliography}
\newpage
\appendices

\section{Dimensions of Object Geometries}
The dimensions of the objects in Fig.~\ref{fig:tool_shapes}, measured from their bounding boxes, are:

\begin{itemize}
    \item \textbf{Mountain}: [0.04 m, 0.04 m, 0.08 m]
    \item \textbf{Rectangle}: [0.05 m, 0.03 m, 0.08 m]
    \item \textbf{Pyramid}: [0.04 m, 0.04 m, 0.08 m]
    \item \textbf{Hexagon}: [0.04 m, 0.035 m, 0.10 m]
    \item \textbf{Cylinder}: [0.04 m, 0.04 m, 0.08 m]
    \item \textbf{Semisphere}: [0.04 m, 0.035 m, 0.117 m]
\end{itemize}


\end{document}